\documentclass[runningheads]{llncs}
\usepackage{graphicx}
\usepackage{subcaption}
\usepackage{xcolor}

\begin{document}

\title{Deep Image Translation for Enhancing Simulated Ultrasound Images}

\titlerunning{Deep Image Translation in Ultrasound Simulation}

\author{Lin Zhang\inst{1} \and
Tiziano Portenier\inst{1} \and
Christoph Paulus\inst{2} \and
Orcun Goksel\inst{1}}

\authorrunning{L. Zhang et al.}
\institute{Computer-assisted Applications in Medicine, ETH Zurich, Switzerland
\and
VirtaMed AG, Schlieren, Switzerland
}
\maketitle              
\begin{abstract}
Ultrasound simulation based on ray tracing enables the synthesis of highly realistic images. 
It can provide an interactive environment for training sonographers as an educational tool. 
However, due to high computational demand, there is a trade-off between image quality and interactivity, potentially leading to sub-optimal results at interactive rates. 
In this work we introduce a deep learning approach based on adversarial training that mitigates this trade-off by improving the quality of simulated images with constant computation time.
An image-to-image translation framework is utilized to translate low quality images into high quality versions.
To incorporate anatomical information potentially lost in low quality images, we additionally provide segmentation maps to image translation.
Furthermore, we propose to leverage information from acoustic attenuation maps to better preserve acoustic shadows and directional artifacts, an invaluable feature for ultrasound image interpretation.
The proposed method yields an improvement of 7.2\% in Fr\'{e}chet Inception Distance and 8.9\% in patch-based Kullback-Leibler divergence.

\end{abstract}

\section{Introduction}
Ultrasound (US) is a low-cost, real-time, and portable diagnostic imaging technique without ionizing radiation, hence widely used in gynecology and obstetrics.
Since its interpretation can be nontrivial due to ultrasound-specific artifacts such as acoustic shadows and tissue-specific speckle texture, sonographer training is crucial.
Ray tracing based US simulation can produce realistic looking images~\cite{mattausch2018realistic}, applicable for ultrasound image simulation in an educational tool.
To that end, the US wavefront is represented with rays on the GPU to simulate interaction with tissue layers, whereas speckle patterns are simulated with a convolutional model of tissue speckle noise.
However, interactive computational constraints often necessitate a compromise in image quality, e.g.\ with limited number of rays or by disabling or reducing essential simulation features. 

Deep learning has achieved great success in various computer vision and graphics tasks. 
In particular, generative adversarial networks (GANs)~\cite{goodfellow2014generative} have been demonstrated as a powerful tool for image synthesis and translation~\cite{isola2017image,zhu2017unpaired}. 
GANs have been widely adapted for various medical image synthesis tasks, such as image inpainting~\cite{armanious2019adversarial} and cross modality translation in both supervised~\cite{armanious2020medgan,nie2018medical} and unsupervised~\cite{zhang2018translating,wolterink2017deep} settings.
In US image synthesis, a two-stage stack GAN was introduced in~\cite{tom2018simulating} for simulating intravascular US imagery conditioned on tissue echogenicity map.
In~\cite{hu2017freehand}, freehand US images are generated conditioned on calibrated physical coordinates. 
Recently in~\cite{vitale2019improving}, feasibility of improving the realism of ray-traced US images has been demonstrated using cycleGAN~\cite{zhu2017unpaired}.

In this work we propose a deep learning based approach for improving the quality of simulated US images that are obtained using a ray tracing algorithm, such that computationally lower-cost (low quality) images can be used to generate higher quality images equivalent to a computationally higher-cost simulation that may not be feasible at interactive frame rates. 
Our framework leverages conditional GANs~\cite{mirza2014conditional} to recover image features that are missing in the low quality images. 
Since both low and high quality images can be simulated using a given algorithm, we tackle this problem in an image-to-image translation setting, where paired low and high quality images are considered to be from different image domains.
Since low quality images may have missing anatomical structures, which introduces ambiguities in the image translation process, we propose to additionally leverage information that is readily available from the underlying simulation algorithm.
For this purpose, we use 2D segmentation map slices within the simulated volume at given transducer locations, that help incorporate all anatomical structures within the imaging field-of-view.
Since major acoustic effects such as shadows are integral and hence global in nature and thus require large network receptive fields reducing capacity, we additionally propose to incorporate easily-computed attenuation maps as additional input to the network.
We show that our proposed methods enhance the final image quality.

\section{Methods}
\subsection{Data Generation}
Simulated US images are generated using a Monte-Carlo ray tracing framework to render B-mode images from a custom geometric fetal model for obstetric training.
US wave interactions are simulated using a surface ray tracing model to find the ray segments between tissue boundaries.
Tissue properties such as acoustic impedance, attenuation, and speed-of-sound are assigned to each tissue type from literature and based on sonographers' feedback form visual inspection.
Along each extracted ray segment, a ray-marching algorithm is applied on the GPU to emulate US scatterer texture by convolving a locally changing point-spread-function with an underlying tissue scatterer representation generated randomly using Gaussian distributions per tissue type~\cite{mattausch2015scatterer}.
Simulated RF data is post-processed with envelope detection, time-gain compensation, log compression, and scan-conversion into Cartesian coordinates, yielding the final gray-scale B-mode images.

\begin{figure}[t]
\minipage{0.24\textwidth}
 \centering
 \includegraphics[width=1\linewidth]{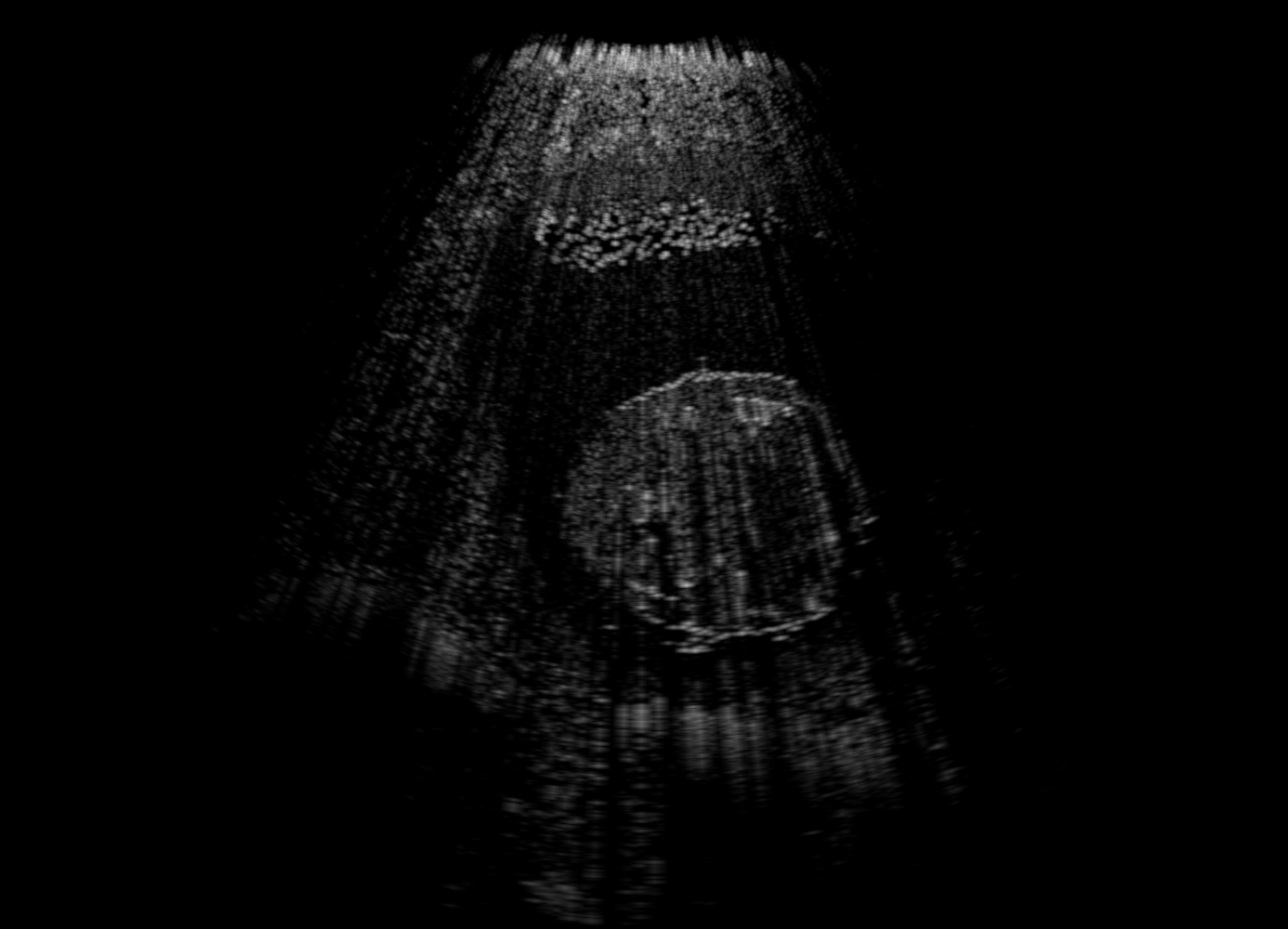}
\endminipage
\hfill
\minipage{0.24\textwidth}
 \centering
 \includegraphics[width=1\linewidth]{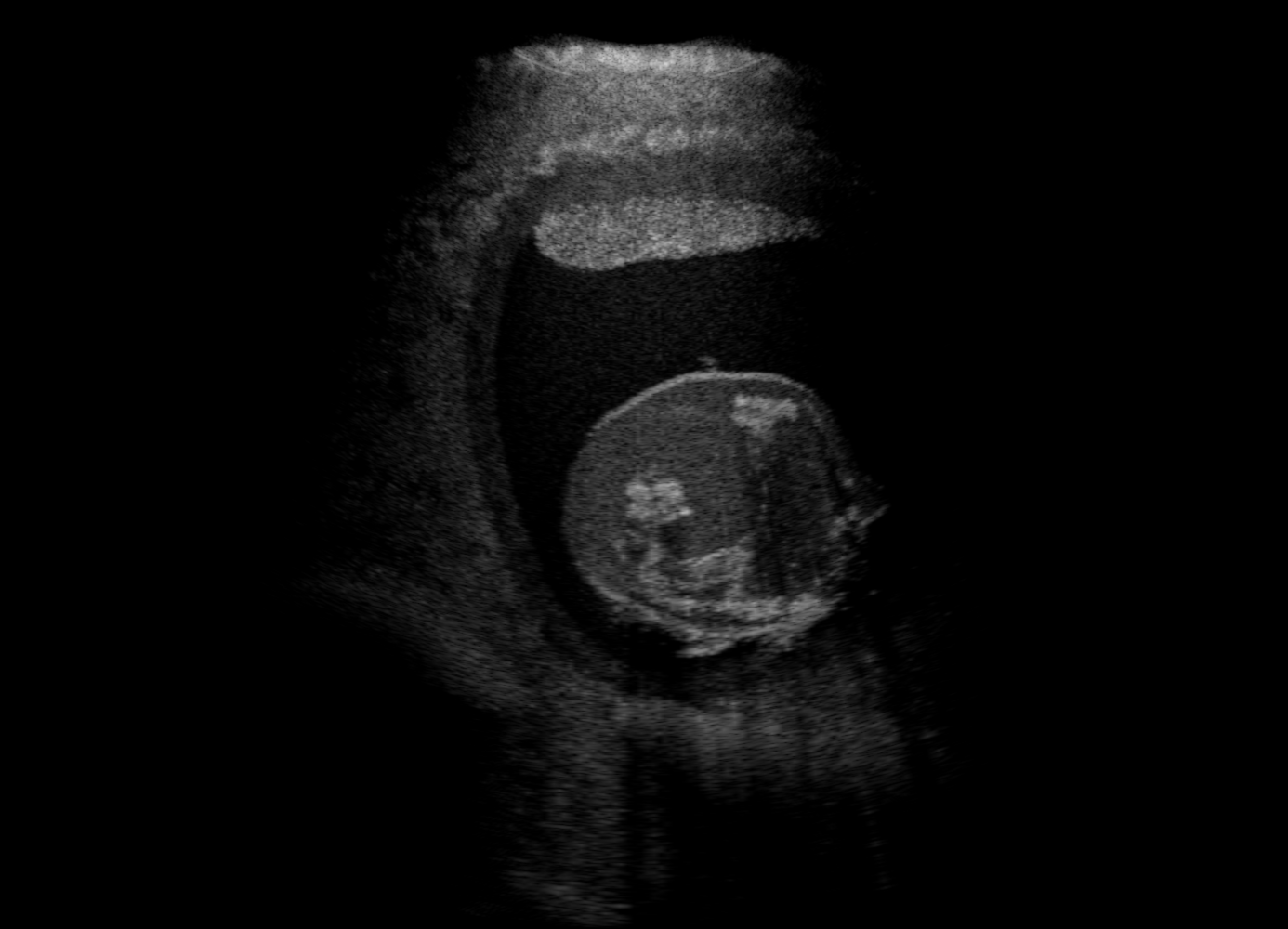}
\endminipage
\hfill
\minipage{0.24\textwidth}
 \centering
 \includegraphics[width=1\linewidth]{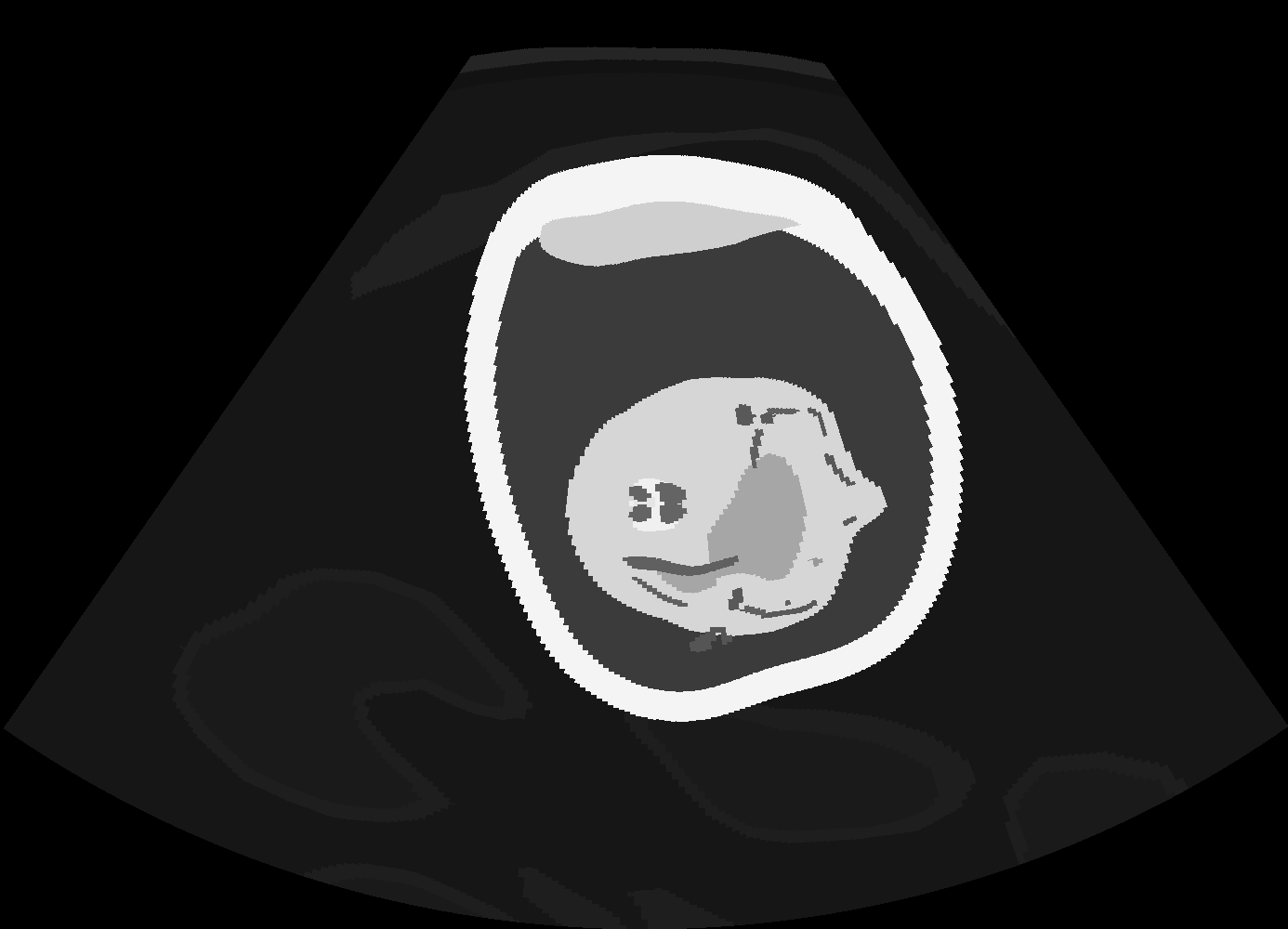}
\endminipage
\hfill
\minipage{0.24\textwidth}
 \centering
 \includegraphics[width=1\linewidth]{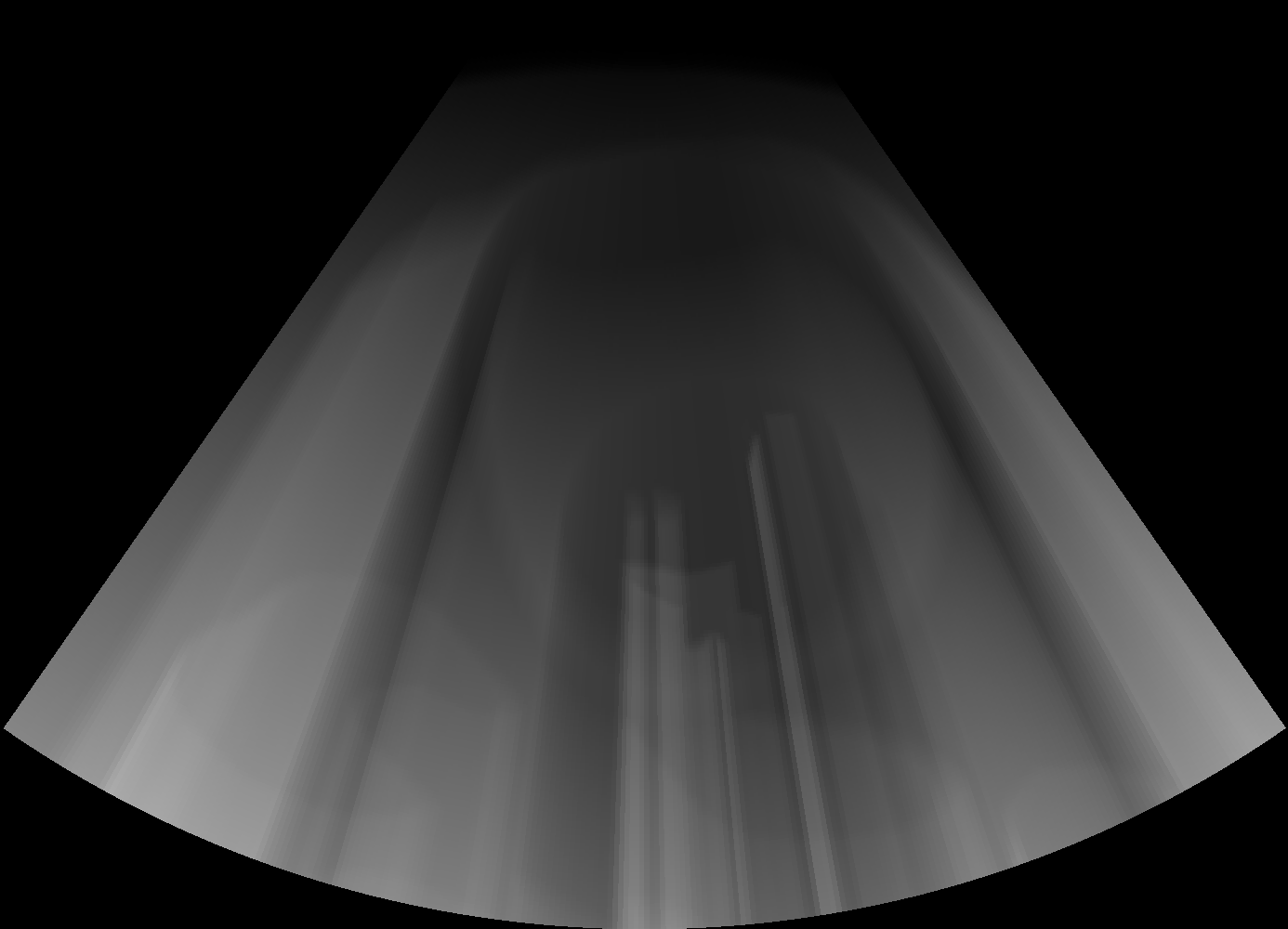}
\endminipage
\hfill
\minipage{0.24\textwidth}
 \centering
 \includegraphics[width=1\linewidth]{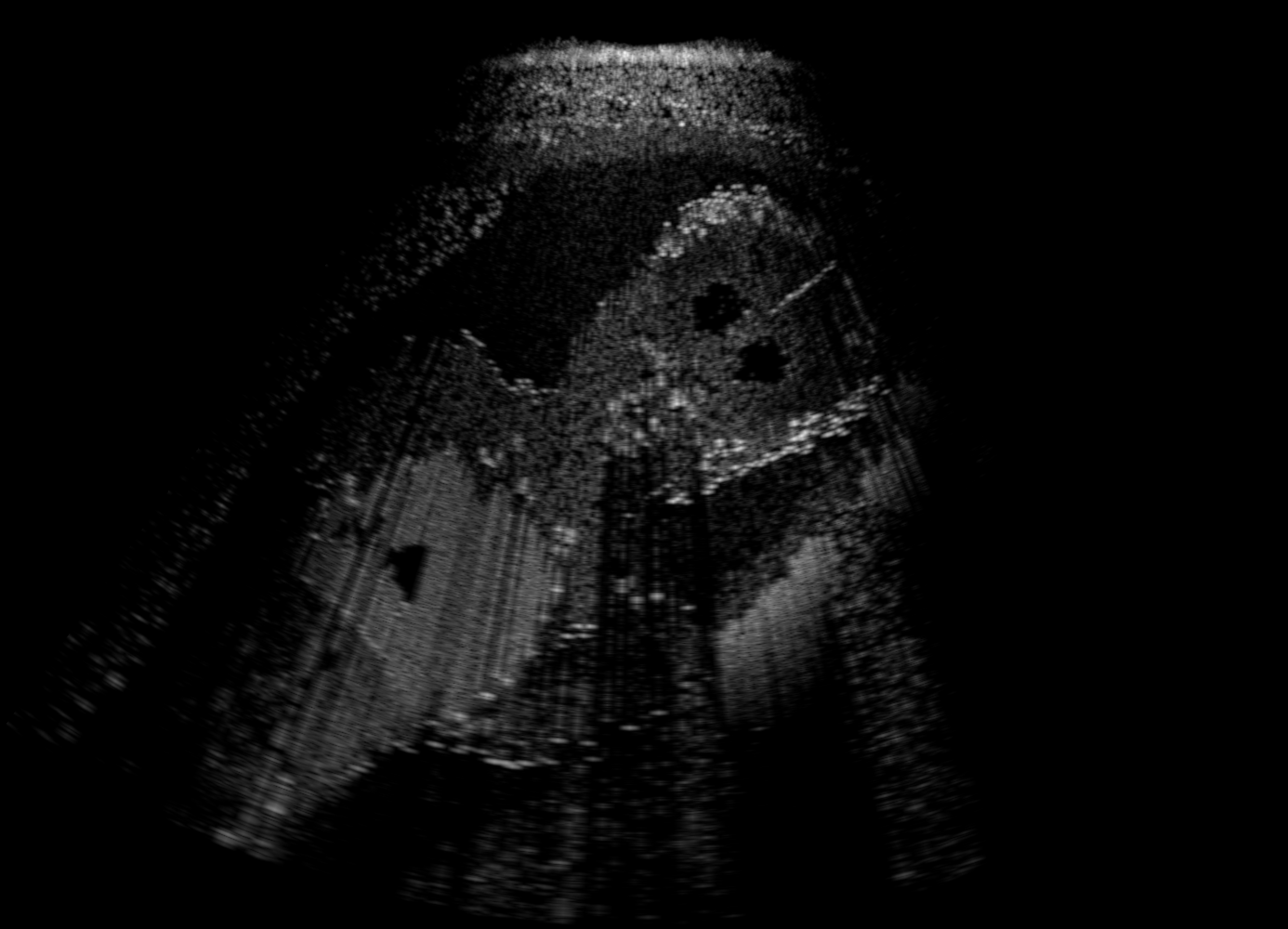}
 \vspace{-2em}\caption*{(a) LQ image}
\endminipage
\hfill
\minipage{0.24\textwidth}
 \centering
 \includegraphics[width=1\linewidth]{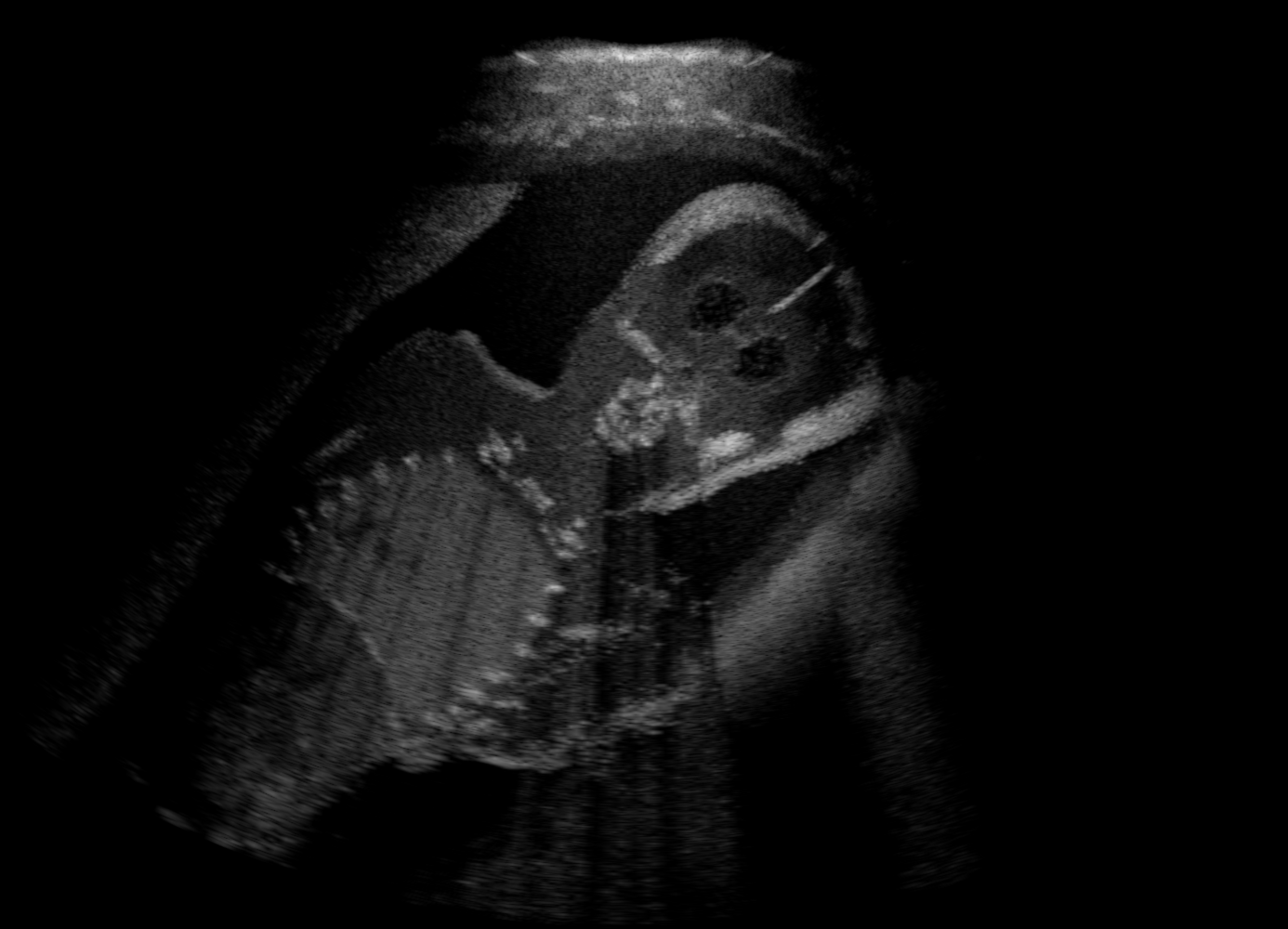}
  \vspace{-2em}\caption*{(b) HQ image}
\endminipage
\hfill
\minipage{0.24\textwidth}
 \centering
 \includegraphics[width=1\linewidth]{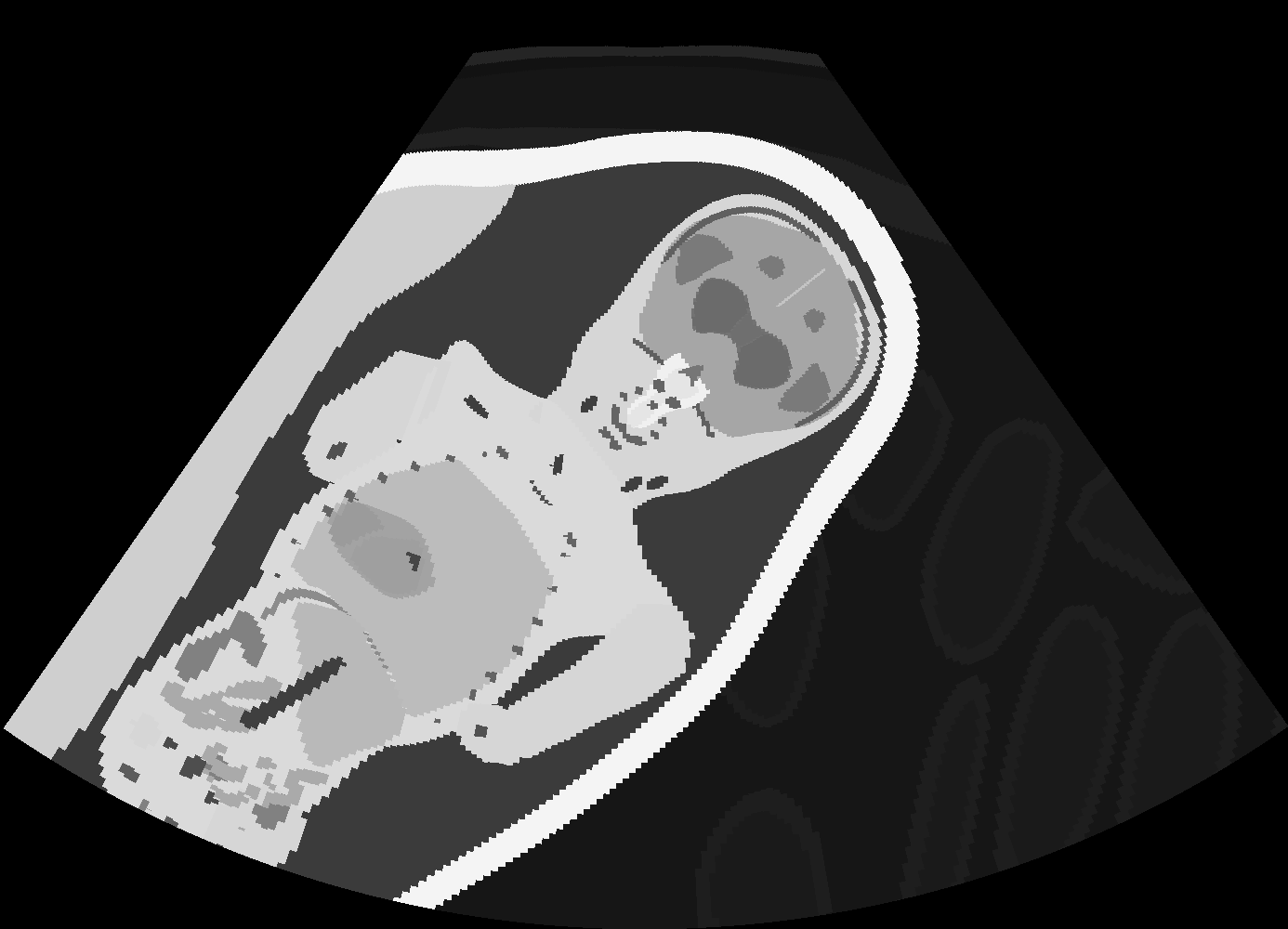}
  \vspace{-2em}\caption*{(c) Segmentation}
\endminipage
\hfill
\minipage{0.24\textwidth}
 \centering
 \includegraphics[width=1\linewidth]{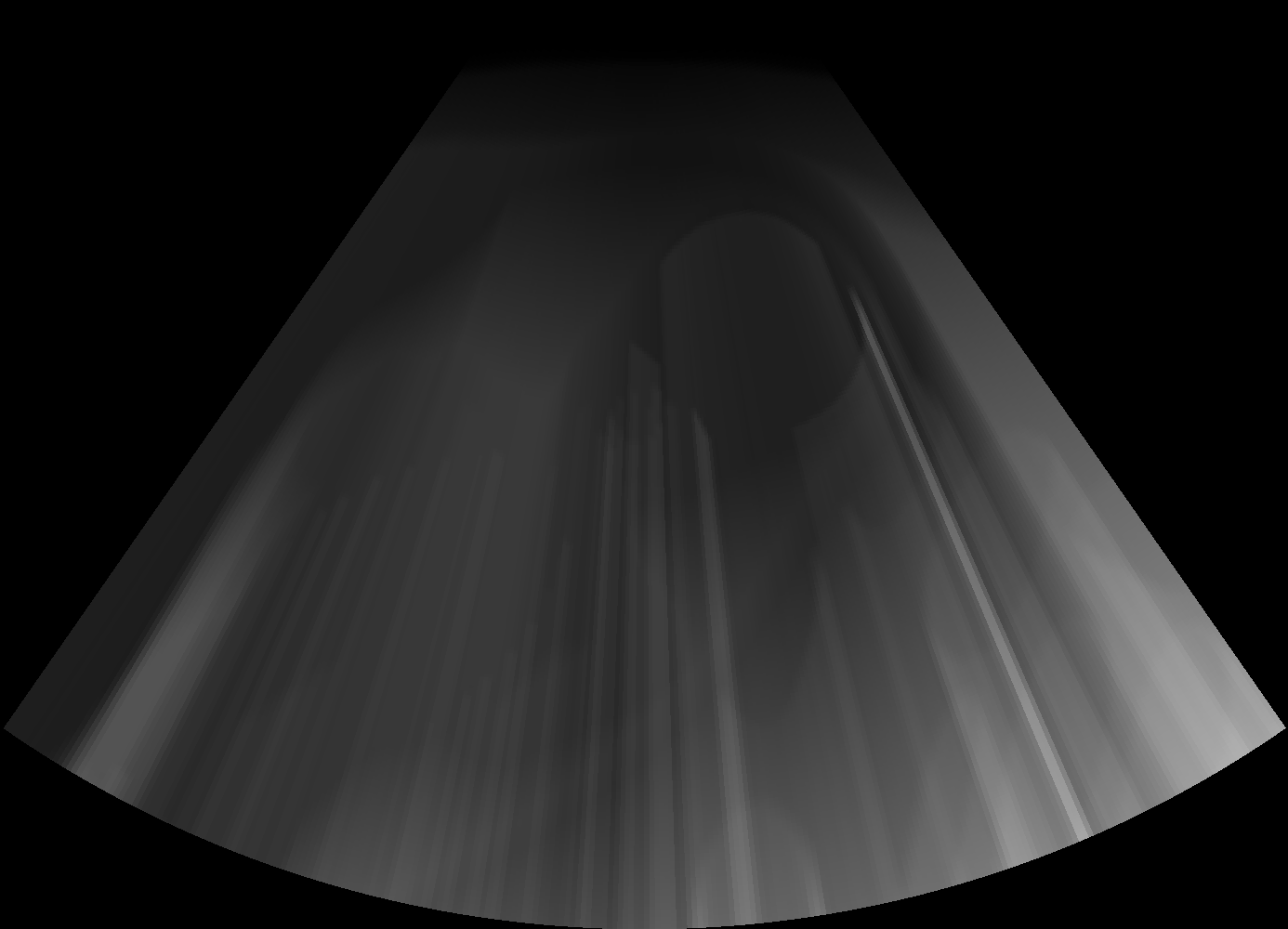}
  \vspace{-2em}\caption*{(d) Attenuation}
\endminipage
\caption{Low quality (a) and high quality (b) simulation outputs, with corresponding segmentation map (c) and integral attenuation map (d).} 
\label{fig:data}
\end{figure}

\vspace{1ex}\noindent{\bf US images. }
For each regularly-sampled key frame of a simulated US fetal exam, paired low and high quality images are generated using two simulation passes: low quality images using one primary ray per US scanline and one elevational layer; and high quality images using 32 primary rays per scanline and three elevational layers~\cite{mattausch2018realistic}. 
Other simulation parameters are kept identical for both simulation passes, cf\ Table~\ref{tab:sim_params}. 
Example B-mode images are shown in Fig.~\ref{fig:data}(a-b).

\begin{table}[b]
\caption{Simulation parameters}
\label{tab:sim_params}
\minipage{0.5\textwidth}
\centering
\begin{tabular}{l||r}
parameter & value \\
\hline 
Triangles fetus & 400k\\
Triangles mother & 275k\\
Image depth & 15.0 cm
\end{tabular}
\endminipage
\minipage{0.5\textwidth}
\centering
\begin{tabular}{l||r}
parameter & value \\
\hline 
Transducer frequency & 8 MHz\\
Transducer field-of-view & 70$^{\circ}$\\
Axial samples & 3072
\end{tabular}
\endminipage
\end{table}

\vspace{1ex}\noindent{\bf Image Mask}
A fixed binary image mask demarcating the imaging region after scan-conversion for the convex probe is also provided as input to the network, in order to constrain the meaningful image translation region and help to save generator capacity.

\vspace{1ex}\noindent{\bf Segmentation Maps}
As additional input for our method, segmentation maps as the cross-section of input triangulated anatomical surfaces are also output by the simulation, corresponding to each low-/high-quality image, cf\ Fig.~\ref{fig:data}(c).

\vspace{1ex}\noindent{\bf Attenuation Maps}
A characteristic feature in real US images is the presence of directional artifacts, which is also valuable for the interpretation of images, for instance in diagnosis of pathology. 
It is therefore important to accurately simulate such artifacts for training purposes. 
Besides reflection and refraction effects, a major source of directional US artifacts is attenuation, which is caused by a reduction in acoustic intensity along the wave travel path due to local tissue effects such as absorption, scattering, and mode conversion. 
Since such artifacts are not only a function of local tissue properties but an integral function along the viewing direction, we propose to directly provide this integrated information to the translation network, hypothesized to improve the quality of translation.

Acoustic intensity arriving at a depth $z$ can be modeled as $I(z)=I_0 e^{-\mu z}$, where $\mu$ is the attenuation constant at a given imaging frequency and $I_0$ is the initial intensity.
Given that the waves travel through different tissue layers with varying attenuation constants $\mu(z)$, the total intensity arriving at a point $z$ can be approximated by
\begin{equation}
    I(z,\mu|_0^z)=I_0 \prod_{i=0}^z e^{-\mu[i]}=I_0 e^{-\sum_{i=0}^z \mu[i]}.
\end{equation}
To approximate such attenuation effect, we create attenuation integral maps $a=e^{-\sum_{i=0}^z \mu[i]}$, accumulated for each image point along the respective ultrasound propagation path.
For better dynamic range and to avoid outliers, these maps are normalized by the 98\,\%ile of image intensities and then scan-converted into the same Cartesian coordinate frame as the simulated B-mode images. 
Fig.~\ref{fig:data}(d) shows sample integral attenuation maps.

\subsection{Image Translation Network}
Our image-to-image translation framework is based on the \emph{pix2pix} network proposed in~\cite{isola2017image}.
Simulated low and high quality US images are considered as source and target domains, respectively, where a translation network $G$ learns a mapping from the source to the target domain. 
Specifically, $G$ maps the low quality US image $x$, the binary mask $m$, the segmentation map $s$, and the attenuation integral map $a$ to the high quality US image $y$, i.e.: $G: \{x,m,s,a\}\rightarrow\{y\}$. 
The discriminator network $D$ is trained to distinguish between real and fake high quality images, conditioned on the corresponding inputs to the generator.
The objective function of the conditional GAN consists of a weighted sum between a GAN loss $L_\mathrm{GAN}$ and a data fidelity term $L_\mathrm{F}$, i.e., 
\begin{eqnarray}
    L &=& L_\mathrm{GAN}(G,D) + \lambda L_\mathrm{F} (G),\\
    L_\mathrm{GAN} &=& \mathbf{E}_{\tilde{x},y}[\log D(y|\tilde{x})] + \mathbf{E}_{\tilde{x}}[\log (1- D(G(\tilde{x})|\tilde{x})],\\
    L_\mathrm{F} &=& \mathbf{E}_{\tilde{x},y}[||y-G(\tilde{x})||_1],
\end{eqnarray}
where $\tilde{x} = (x,m,s,a)$.
Before computing the losses, the output is element-wise multiplied with the binary mask to restrict the loss to the relevant output regions. 

\begin{figure}
\centering
\includegraphics[width=0.9\textwidth]{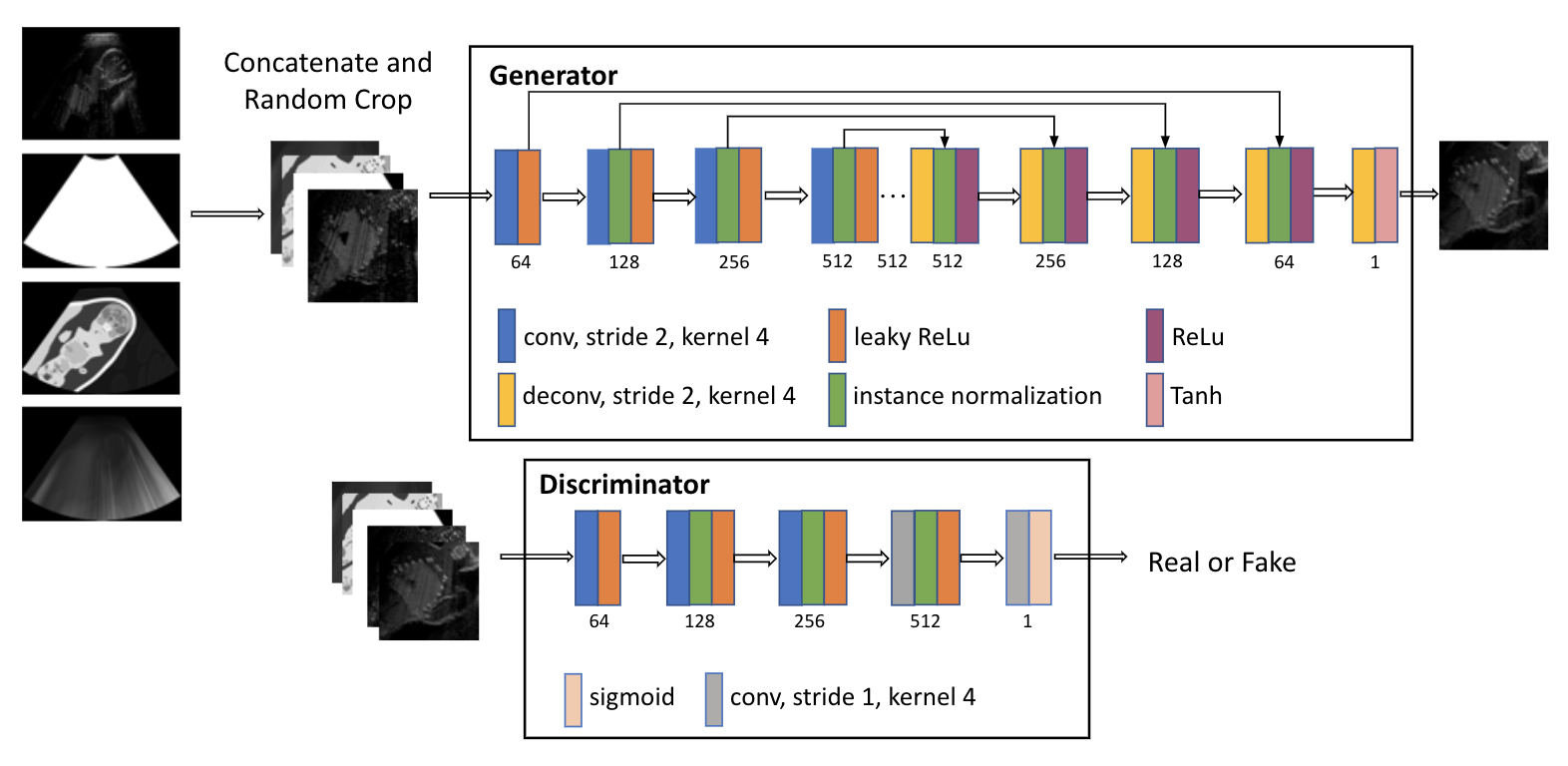}
\caption{Network Architecture}
\label{fig:network}
\end{figure}

As proposed in~\cite{isola2017image}, we parametrize $G$ using a 8-layer Unet with skip connections and $D$ using a 4-layer convolutional network, i.e.\ a \emph{patchGAN} discriminator. 
Instance normalization is applied before nonlinear activation.
The full field-of-view B-mode images from the simulation are of size $1000\times1386$ pixels.
Applying pix2pix directly at such high resolution may lead to unsatisfactory results, as reported in~\cite{wang2018high}. 
We therefore use randomly cropped patches of a smaller size.
A patch size of $512\times512$ pixels is found empirically to provide sufficient anatomical context, without degradation in image quality.
Fig.~\ref{fig:network} shows an overview of our network architecture. 

\section{Experiments and Results}
\vspace{1ex}\noindent{\bf Implementation Details and Network Training. }
We use the Adam optimizer~\cite{kingma2014adam} with a learning rate of 0.0002 and exponential decay rates $\beta_1=0.5$ and $\beta_2=0.999$.
Since with GANs the Nash equilibrium is often not reached in practice, we early stop training at 50k iterations.
We use a batch size of 16 and set $\lambda=100$.
Our dataset consists of 6669 4-tuples $(x, y, s, a)$ and a constant binary mask $m$ covering the beam shape for all samples. 
We use randomly-selected 6000 images for training and the rest for evaluation. 
To quantitatively evaluate our models, from each test image we randomly crop four patches of size $512\times512$, yielding an evaluation set of 2676 image patches that are not seen during training.
Note that our original dataset consists of images that are temporally far apart, thus the test images cannot be temporally consecutive and thus inherently similar to any training images.

\vspace{1ex}\noindent{\bf Comparative Evaluation. }
To demonstrate the effectiveness of the proposed additional inputs from the image formation process, we conduct an ablation study by considering different combinations of network inputs. 
We refer the pix2pix network with low quality image and binary mask in the input channel as our baseline \textit{L2H$_\mathrm{M}$}. 
We compare this baseline with the following variants:
1) \textit{L2H$_\mathrm{MS}$}: \textit{L2H$_\mathrm{M}$} with segmentation map $s$ as additional input;
2) \textit{L2H$_\mathrm{MSA}$}: \textit{L2H$_\mathrm{MS}$} with attenuation integral map $a$ as additional input.

\begin{figure}[t]
\minipage{0.195\textwidth}
 \centering
 \includegraphics[width=1\linewidth]{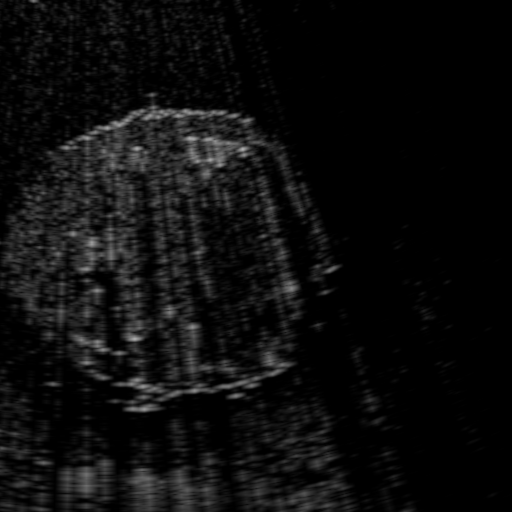}\\
 \includegraphics[width=1\linewidth]{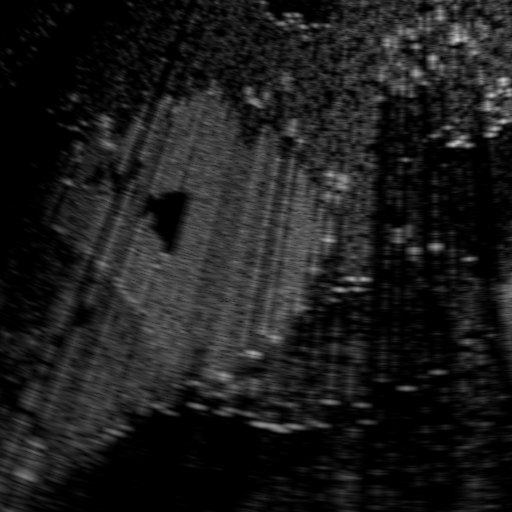} 
 \includegraphics[width=1\linewidth]{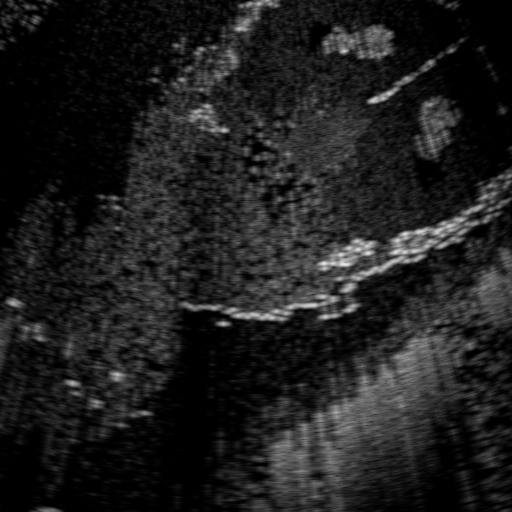} 
 \includegraphics[width=1\linewidth]{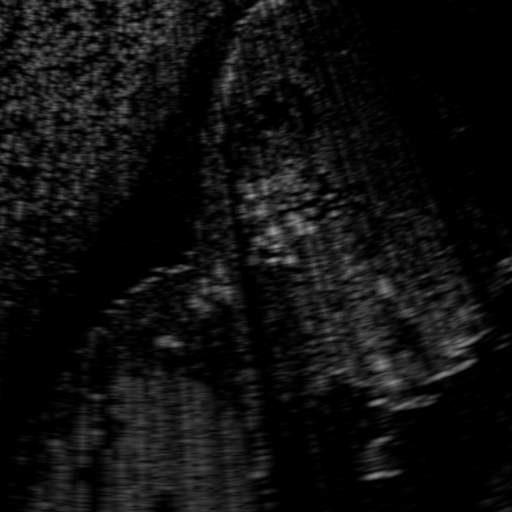} 
 \vspace{-2em}\caption*{(a) Input}
\endminipage
\hfill
\minipage{0.195\textwidth}
 \centering
 \includegraphics[width=1\linewidth]{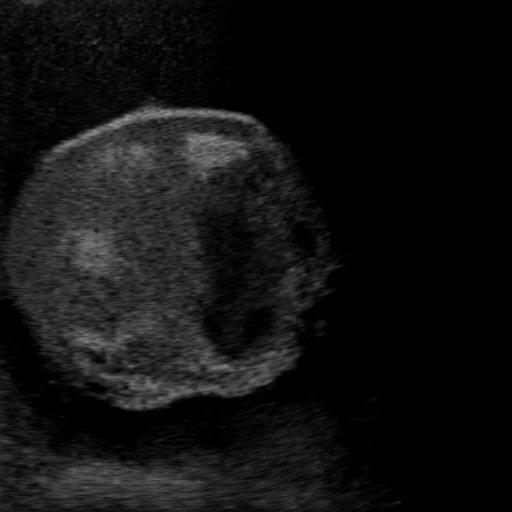}
 \includegraphics[width=1\linewidth]{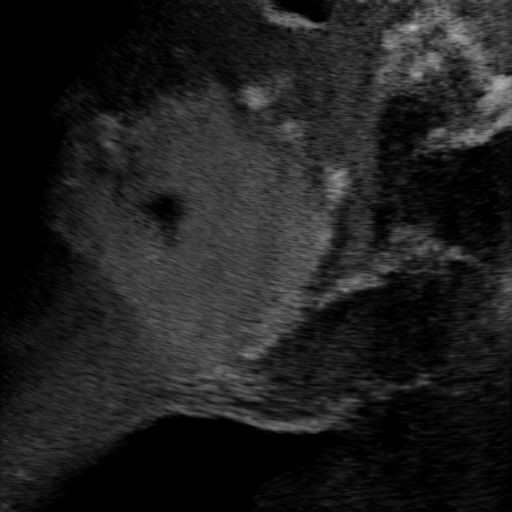} 
 \includegraphics[width=1\linewidth]{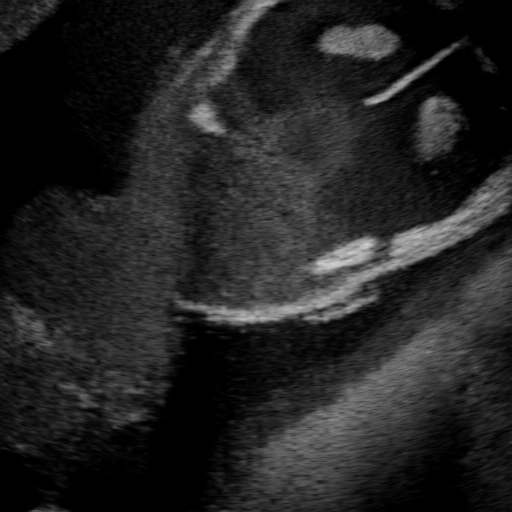} 
\includegraphics[width=1\linewidth]{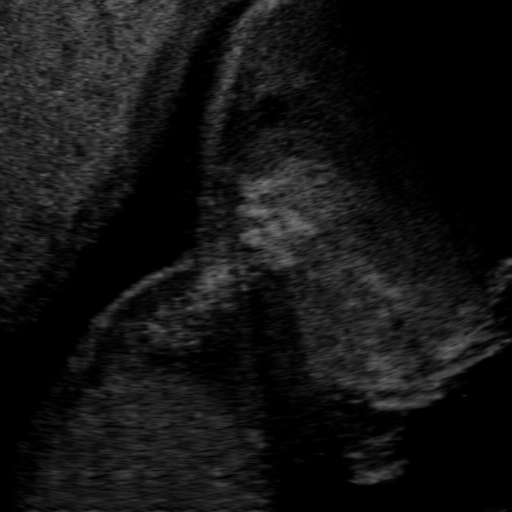} 
 \vspace{-2em}\caption*{(b) L2H$_\mathrm{M}$}
\endminipage
\hfill
\minipage{0.195\textwidth}
 \centering
 \includegraphics[width=1\linewidth]{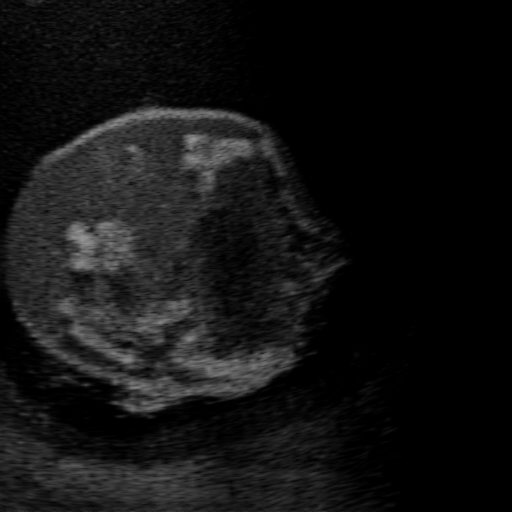}
 \includegraphics[width=1\linewidth]{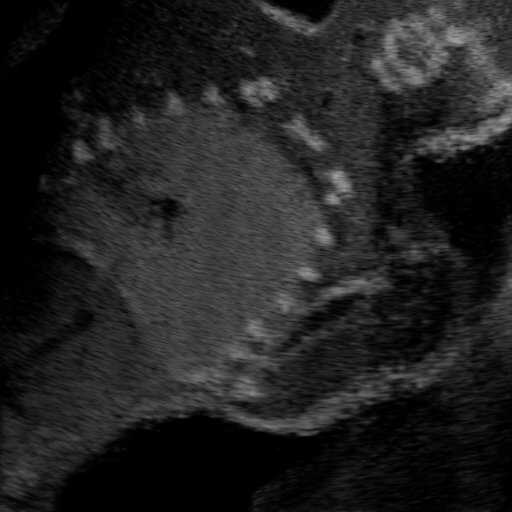} 
 \includegraphics[width=1\linewidth]{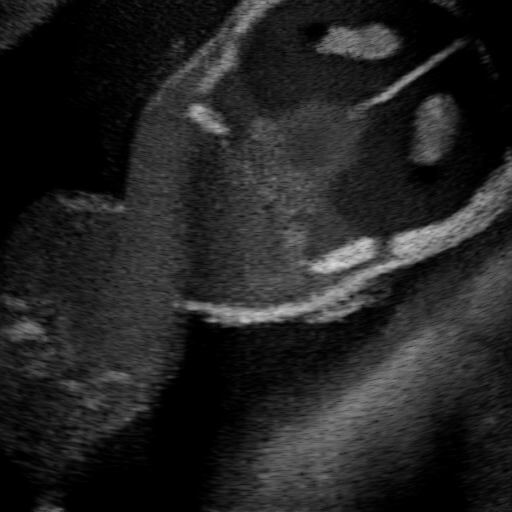} 
 \includegraphics[width=1\linewidth]{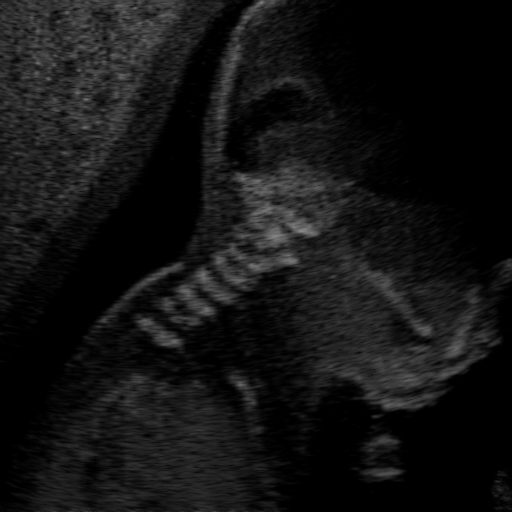} 
 \vspace{-2em}\caption*{(c) L2H$_\mathrm{MS}$}
\endminipage
\hfill
\minipage{0.195\textwidth}
 \centering
 \includegraphics[width=1\linewidth]{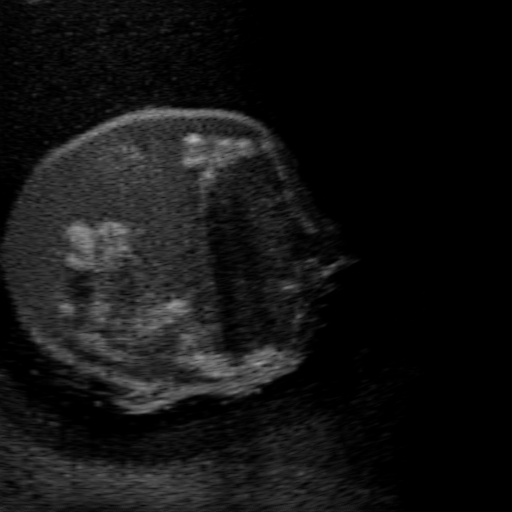}
 \includegraphics[width=1\linewidth]{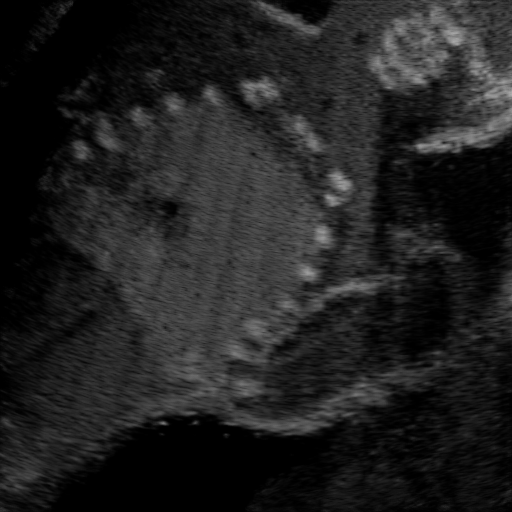} 
 \includegraphics[width=1\linewidth]{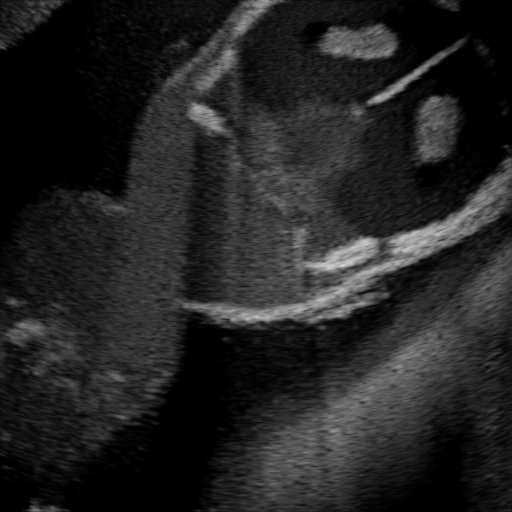} 
 \includegraphics[width=1\linewidth]{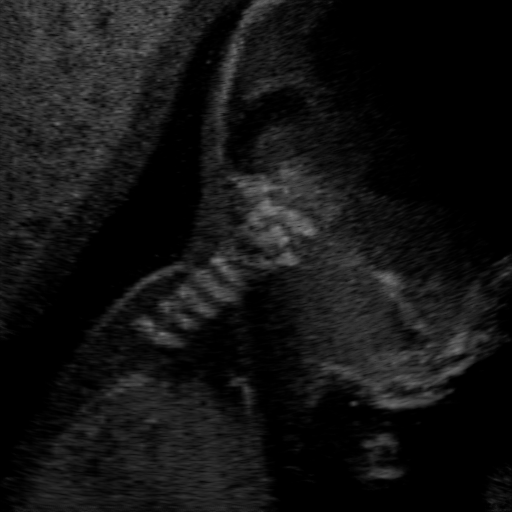} 
 \vspace{-2em}\caption*{(d) L2H$_\mathrm{MSA}$}
\endminipage
\hfill
\minipage{0.195\textwidth}
 \centering
 \includegraphics[width=1\linewidth]{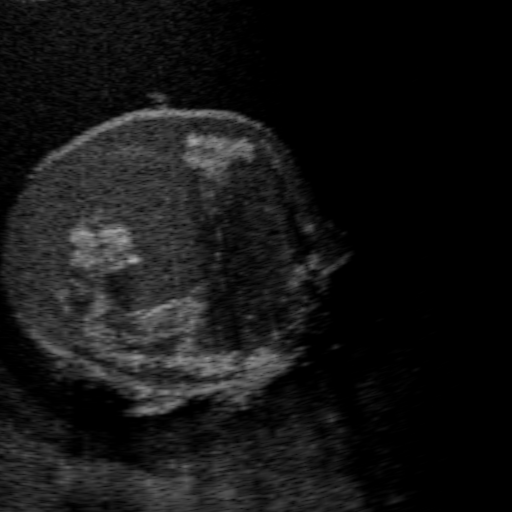}
 \includegraphics[width=1\linewidth]{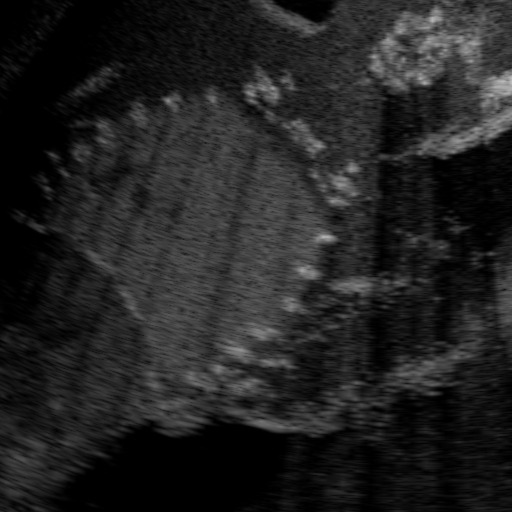} 
 \includegraphics[width=1\linewidth]{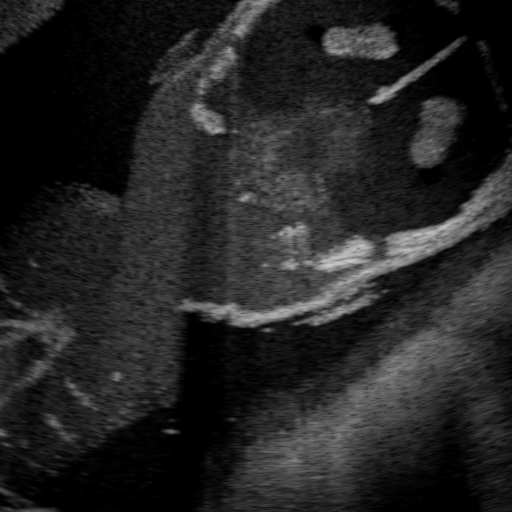} 
 \includegraphics[width=1\linewidth]{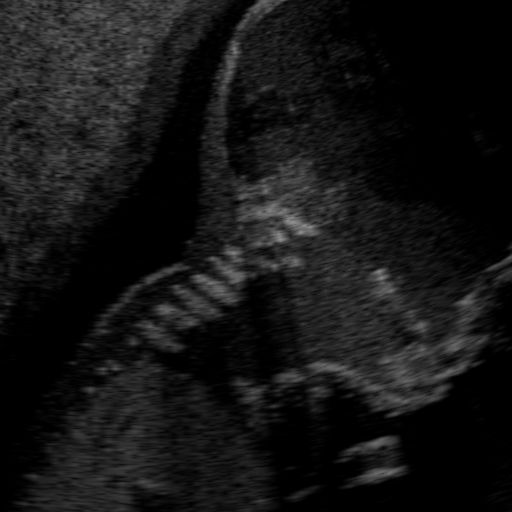} 
 \vspace{-2em}\caption*{(e) Target}
\endminipage
\caption{Low-quality input (a), GAN outputs (b-d), and high-quality target (e).} 
\label{fig:qual_results}
\end{figure}

\vspace{1ex}\noindent{\bf Qualitative Results. }
Fig.~\ref{fig:qual_results} shows a visual comparison of the three model variants on four examples. 
The baseline \textit{L2H$_\mathrm{M}$} fails to preserve anatomical structures due to missing structural information in the input images. 
Resulting ambiguities in the network prediction cause artifacts such as blur in regions that feature fine details such as bones. 
Providing segmentation maps as additional input (\textit{L2H$_\mathrm{MS}$}) greatly reduces such artifacts as shown in Fig.~\ref{fig:qual_results}(c). 
However, \textit{L2H$_\mathrm{MS}$} still struggles in modeling complex non-local features such as directional occlusion artifacts, note the lack of acoustic shadows in Fig.~\ref{fig:qual_results}(c). 
In contrast, our final model \textit{L2H$_\mathrm{MSA}$} is able to accurately synthesize these features and produces translations significantly closer to the target, as demonstrated in Fig.~\ref{fig:qual_results}(d). 
In particular, our proposed model with segmentation and attenuation integral maps is able to recover both missing anatomical structures and directional artefacts.

\vspace{1ex}\noindent{\bf Quantitative Results}
The effectiveness of the proposed model is further evaluated using the following quantitative metrics:

1) PSNR: Peak signal-to-noise ratio between two images $A$ and $B$ is defined by $\textrm{PSNR} = 10\log_{10}(\frac{255}{\textrm{MSE}})$ with mean squared error MSE between $A$ and $B$.

2) SSIM: Structural similarity index quantifies the visual changes in structural information as $\textrm{SSIM}(A, B)$=$\frac{(2\mu_A\mu_B+c_1)(2\sigma_{AB}+c_2)}{(\mu_A^2+\mu_B^2+c_1)(\sigma_A^2+\sigma_B^2+c_2)}$ with regularization constants $c_1$ and $c_2$, local means $\mu_A$ and $\mu_B$, local standard deviations $\sigma_A$ and $\sigma_B$, and cross covariance $\sigma_{AB}$.
We use the default parameters of the MATLAB implementation to compute the metric. 

3) pKL: Speckle appearance, relevant for tissue characterization in US images~\cite{shankar1993use}, affects image histogram statistics.
Hence, discrepancy in histogram statistics can quantify differences in tissue specific speckle patterns.
Kullback Leibler divergence compares normalized histograms $h_A$ and $h_B$ of two images $A$ and $B$ as:
$ \textrm{KL} (h_A || h_{B}) = \sum_{l=1..d} h_A[l] \log\left(\frac{h_A[l]}{h_{B}[l]}\right)$. 
We set the number of histogram bins $d$ to $50$.
To emphasize structural differences, we calculate KL divergence locally within $32\times32$ sized non-overlapping patches and report the metric mean, called \emph{patch KL} (pKL) herein.

4) FID: Fr\'{e}chet Inception Distance compares the distributions of generated samples and real samples by computing the distance between two multivariate Gaussians fitted to hidden activations of Inception network~v3. 
This is a widely used metric to evaluate GAN performance, capturing both perceptual image quality and mode diversity. 
For this purpose, center crops of test images were sub-divided into four pieces of $299\times299$, to match Inception v3 input size.

Tab.~\ref{tab:quant_results} summarizes quantitative results for all models and all metrics, with the additional comparison to the discrepancy between low quality and high quality images as reference.
A preliminary baseline experiment without GAN loss resulted in very blurry images with an FID score of 184.71.
The results in Tab.~\ref{tab:quant_results} demonstrate that \textit{L2H$_\mathrm{MSA}$} achieves the best translation performance in terms of all proposed metrics.
The effectiveness of providing informative inputs to the network is well demonstrated in the gradual improvement in PSNR, SSIM and pKL, showing higher fidelity in anatomical structures and directional shadow artifacts.
The metric pKL gives further indication of closer speckle appearance achieved by \textit{L2H$_\mathrm{MSA}$}.
Moreover, FID score indicates higher statistical similarity between the target and generated images using the proposed final model, with an improvement of $7.2\%$ compared to the baseline (\textit{L2H$_\mathrm{M}$}). 

\begin{table}[b]
\setlength{\tabcolsep}{2pt}
\caption{Quantitative results. \%ile refers to 5 percentile values for PSNR and SSIM and 95 percentile otherwise. Bold number indicates the best performance.}
\centering
   \begin{tabular}{l|rr|rr|rr|r}
      & \multicolumn{2}{c|}{PSNR} &
      \multicolumn{2}{c|}{SSIM [\%]} &\multicolumn{2}{c|}{pKL ($\times 10^{2}$)} &\multicolumn{1}{c}{FID}\\
 & mean & \%ile & mean & \%ile & mean & \%ile & \\
\hline
Low quality & 25.31 & 20.18  & 64.05 & 35.10  &38.90 &82.02 &204.60 \\
L2H$_\mathrm{M}$  & 29.07 & 24.62  & 70.75 & 45.73  &15.14 &31.45 &17.88 \\
L2H$_\mathrm{MS}$ & 29.26 & 24.78 & 71.22 & 46.37  & 14.57 & 31.41 &17.62 \\
L2H$_\mathrm{MSA}$  & \bf{29.40} & \bf{24.89} & \bf{71.47} & \bf{46.67}  & \bf{13.80} & \bf{29.02} &\bf{16.59} 
\end{tabular}
\label{tab:quant_results}
\end{table}

\subsection{Full field-of-view Images}
\begin{figure}[t]
\minipage{0.195\textwidth}
 \centering
 \includegraphics[width=1\linewidth]{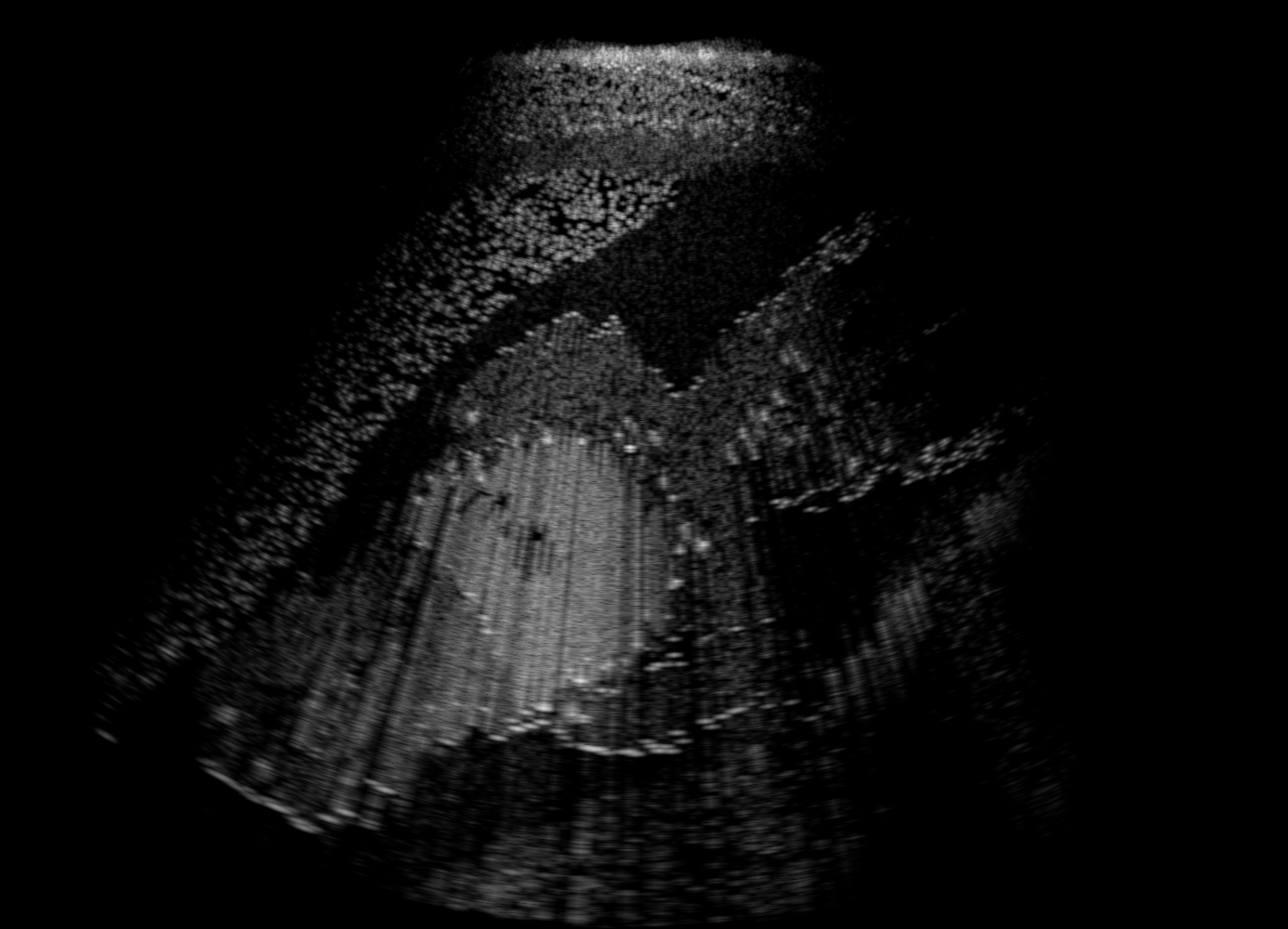}
 \includegraphics[width=1\linewidth]{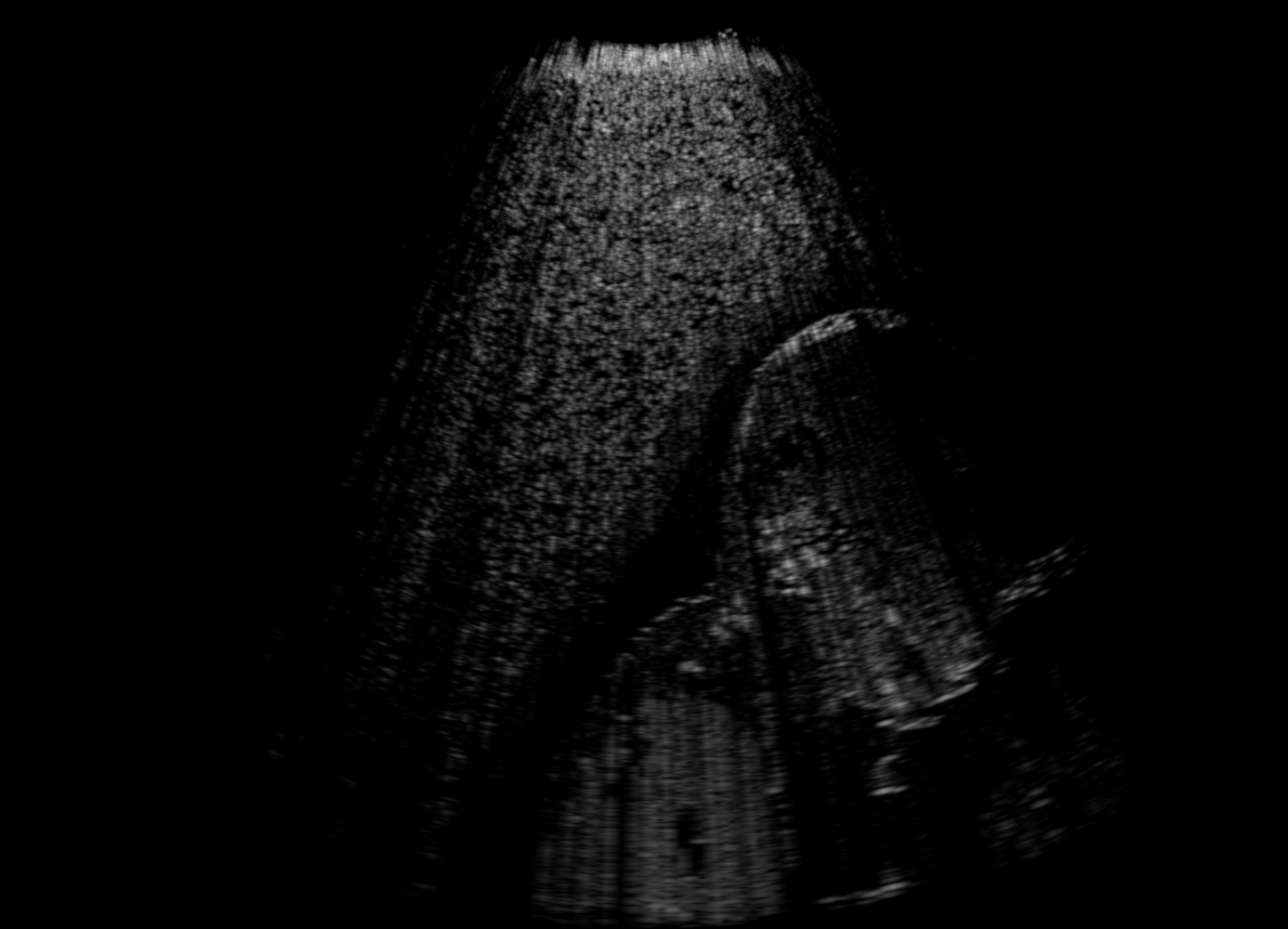}
 (a) Input
\endminipage
\hfill
\minipage{0.195\textwidth}
 \centering
 \includegraphics[width=1\linewidth]{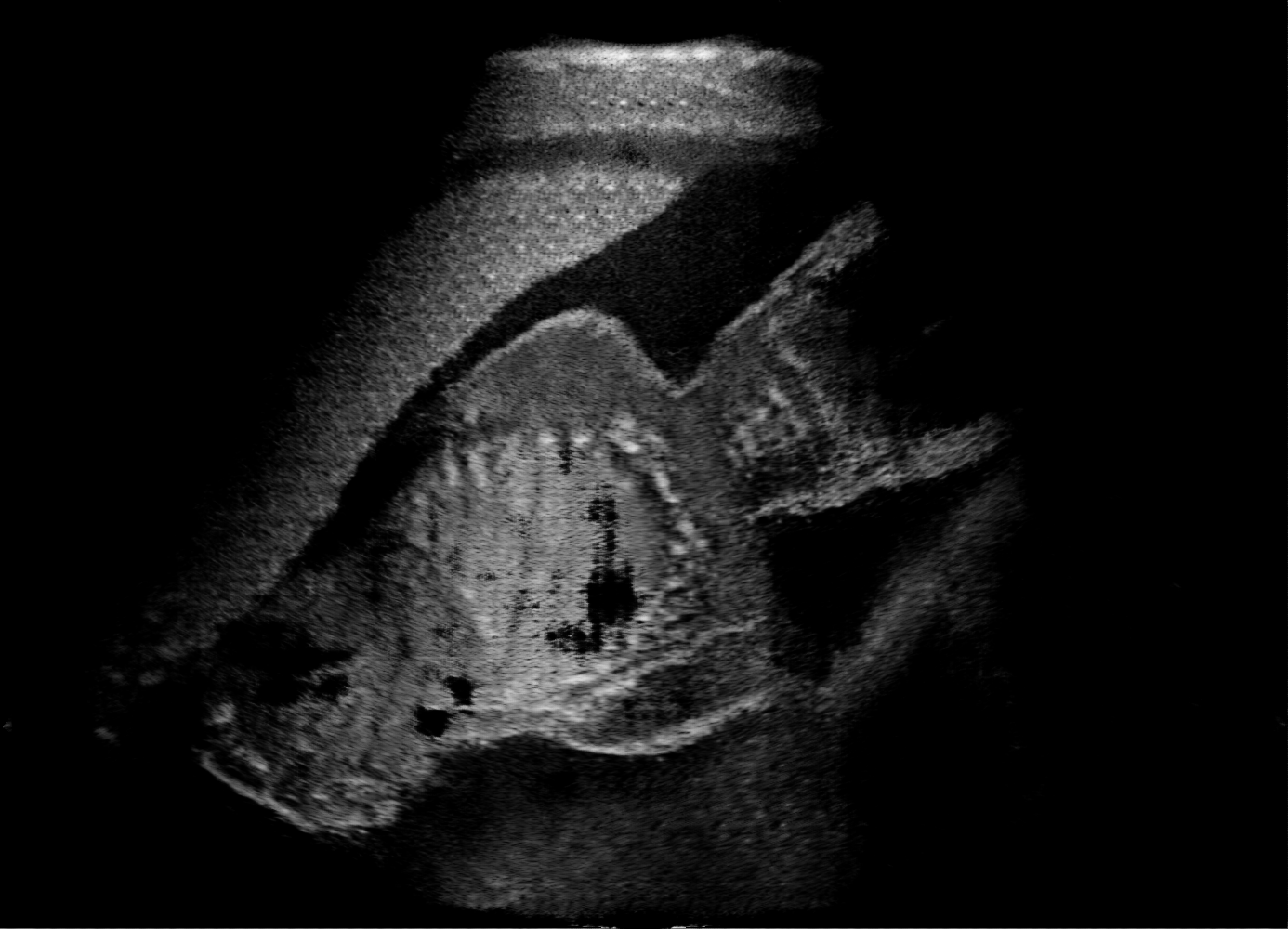}
 \includegraphics[width=1\linewidth]{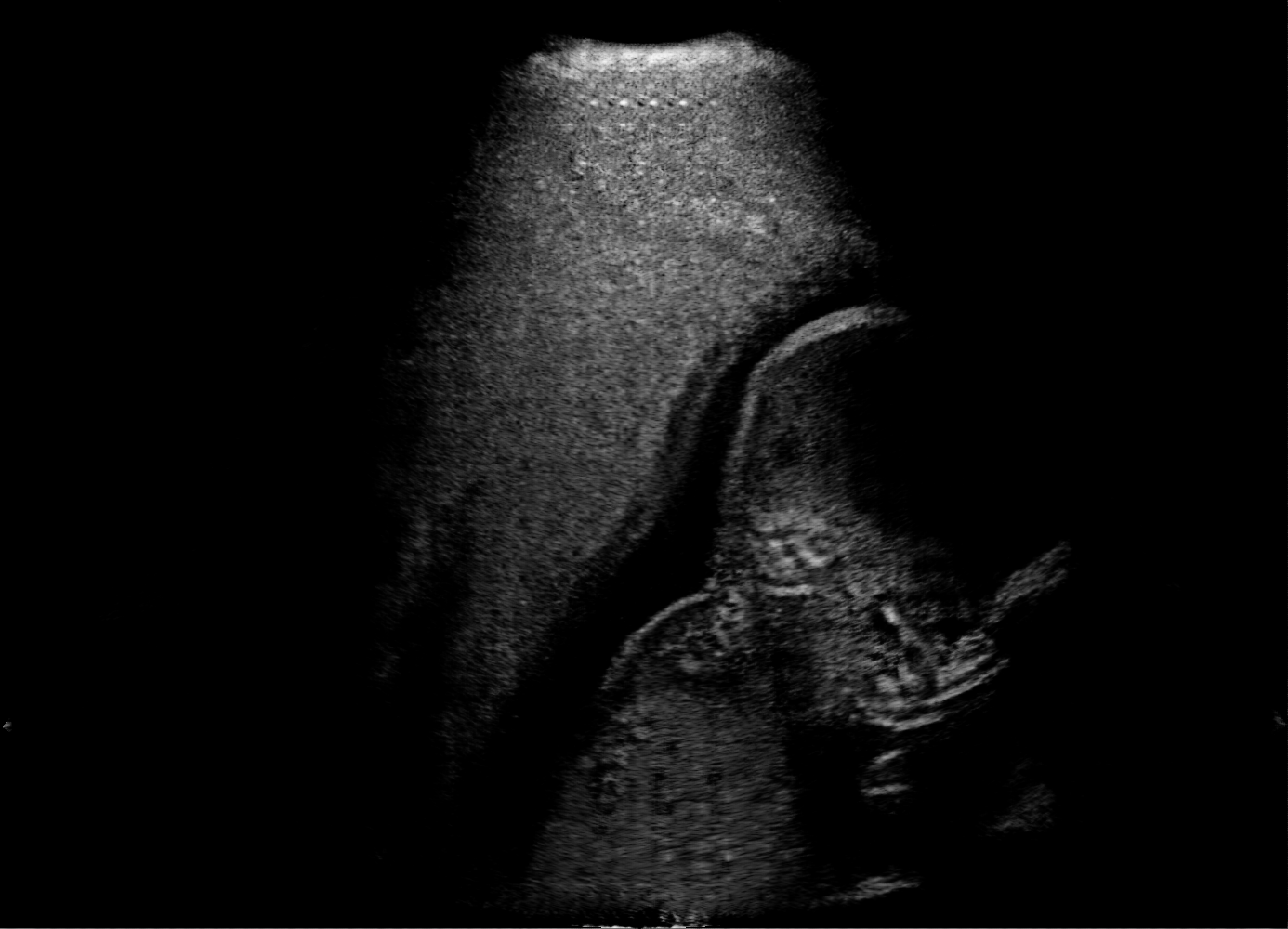}
 (b) L2H$_\mathrm{M}$
\endminipage
\hfill
\minipage{0.195\textwidth}
 \centering
 \includegraphics[width=1\linewidth]{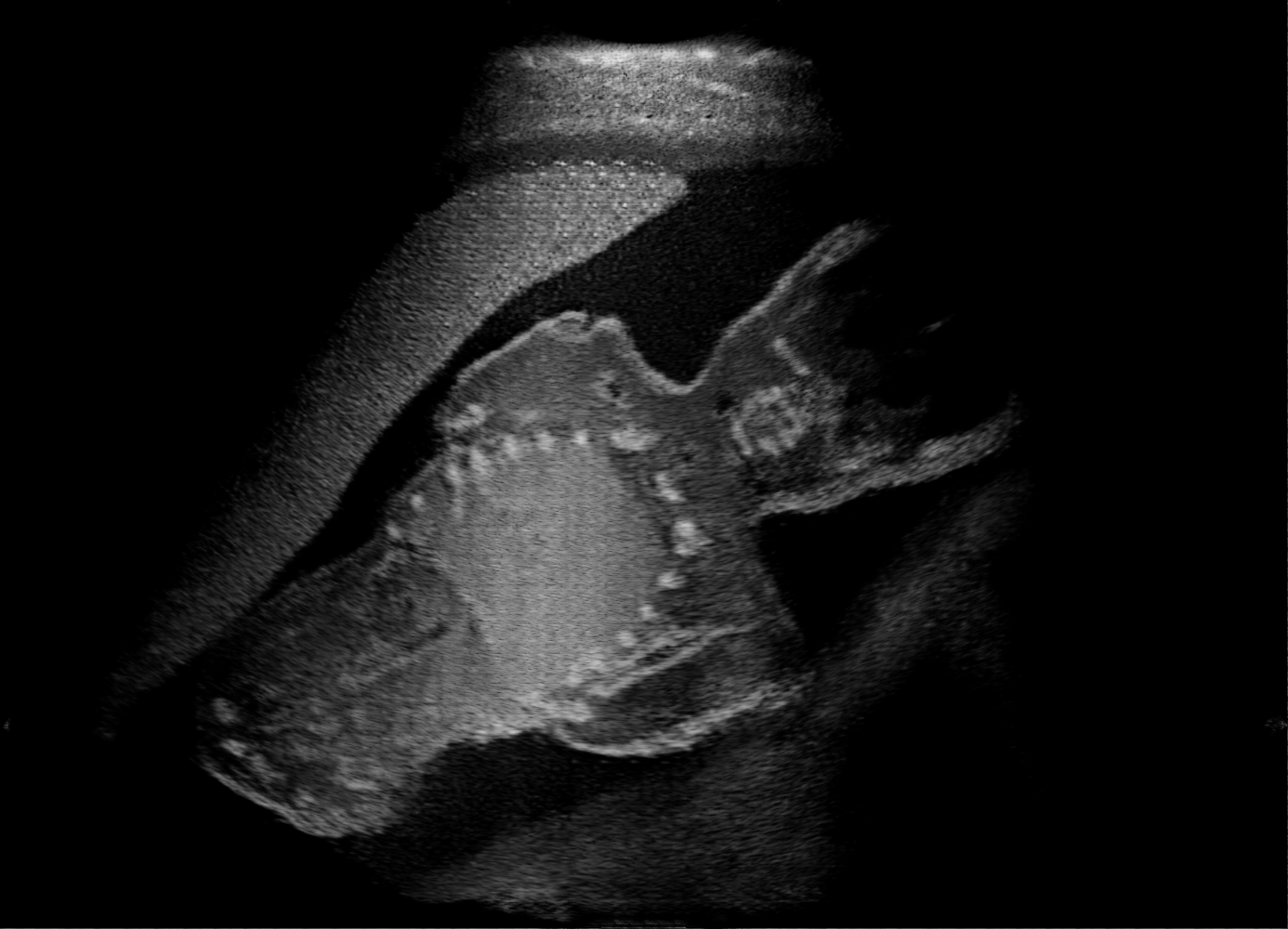}
 \includegraphics[width=1\linewidth]{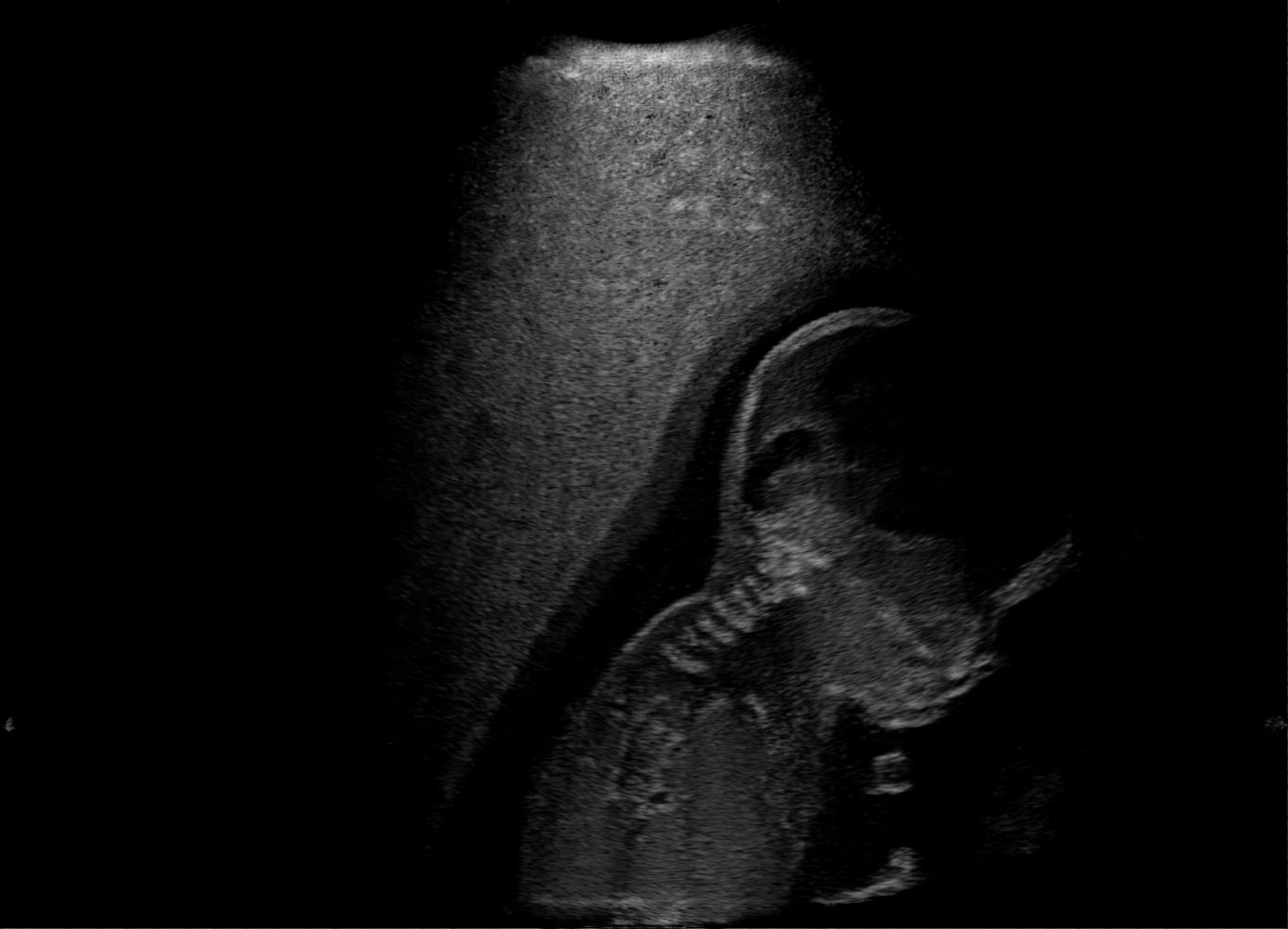}
 (c) L2H$_\mathrm{MS}$
\endminipage
\hfill
\minipage{0.195\textwidth}
 \centering
 \includegraphics[width=1\linewidth]{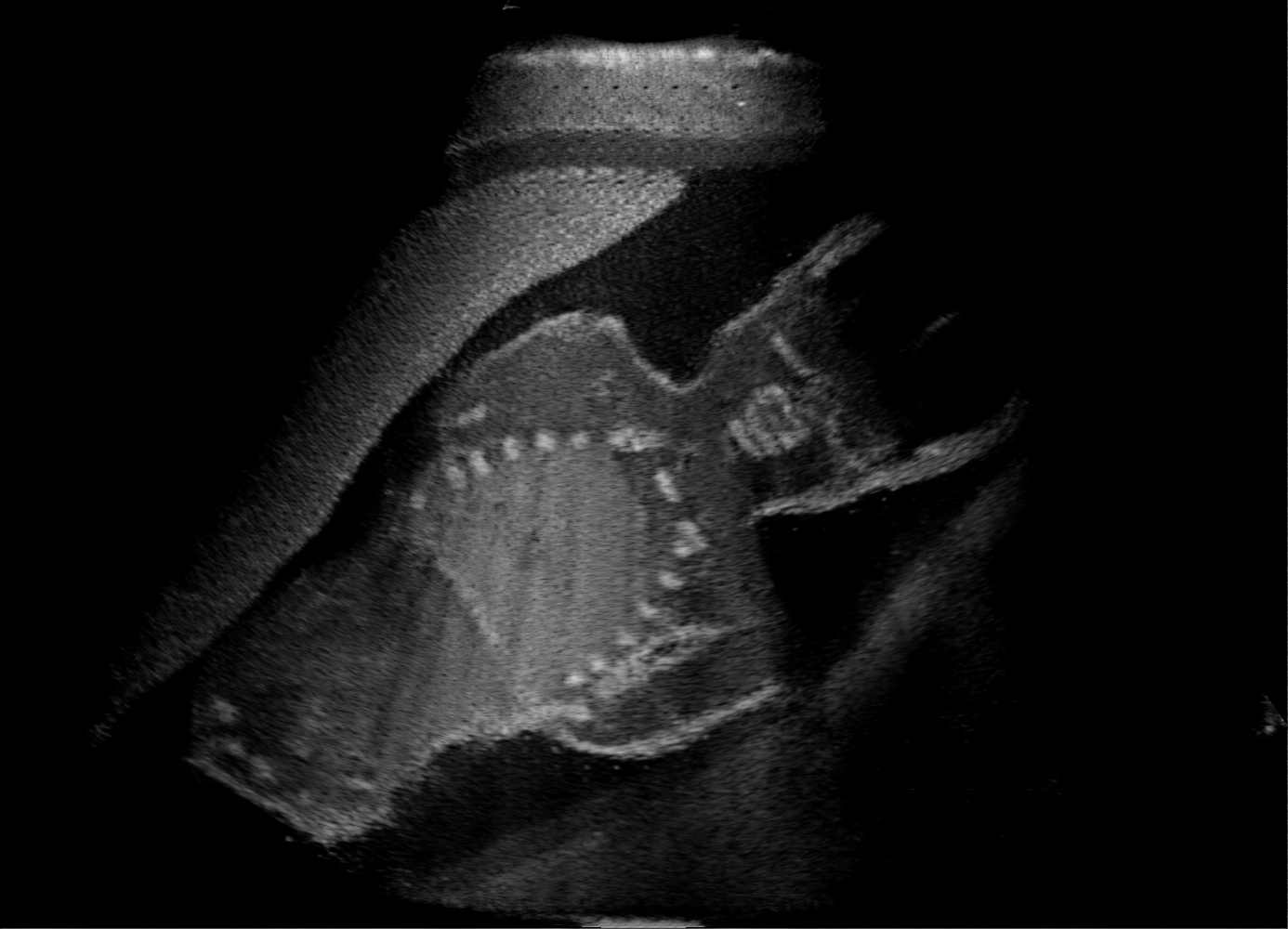}
 \includegraphics[width=1\linewidth]{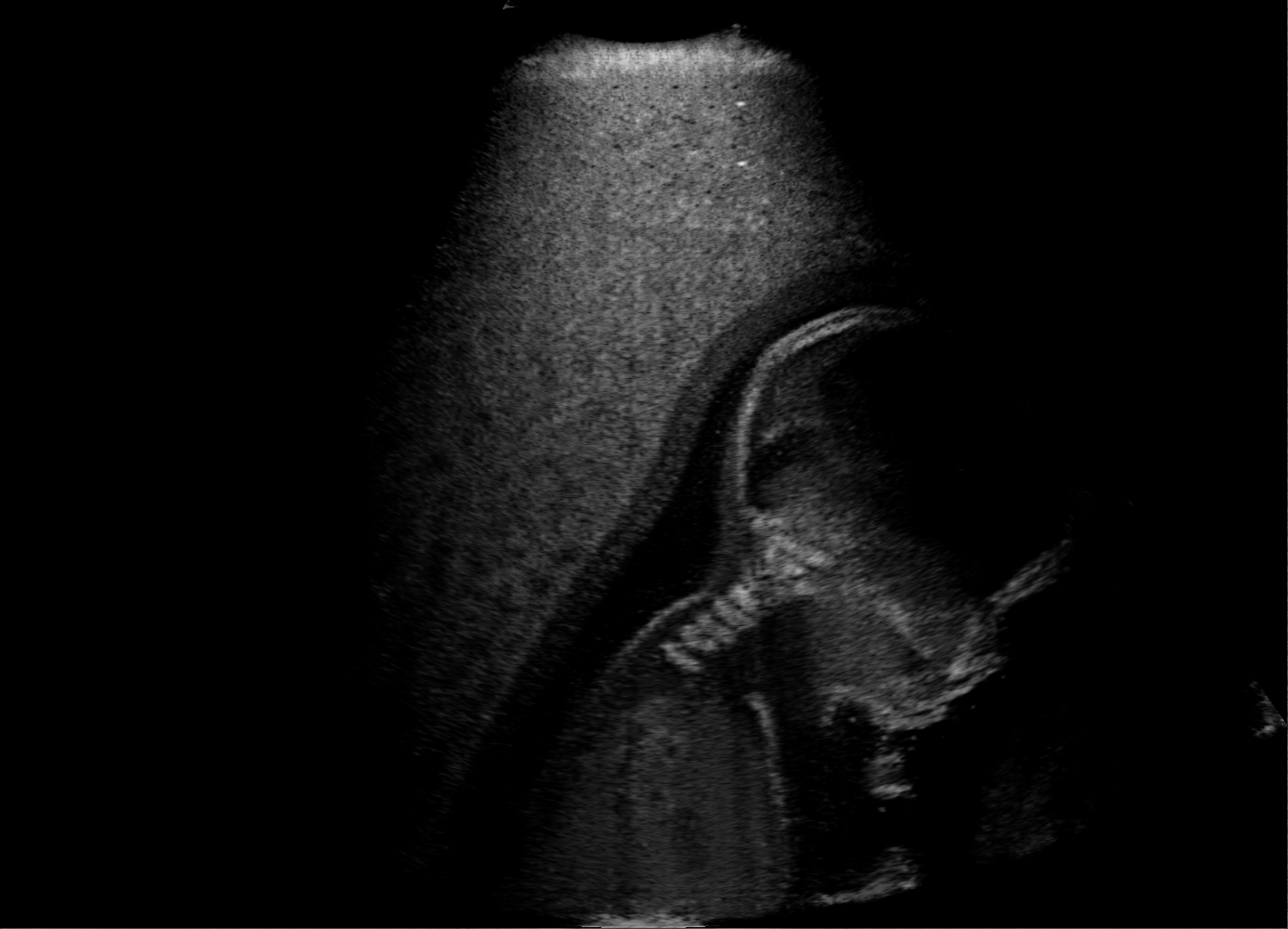}
 (d) L2H$_\mathrm{MSA}$
\endminipage
\hfill
\minipage{0.195\textwidth}
 \centering
 \includegraphics[width=1\linewidth]{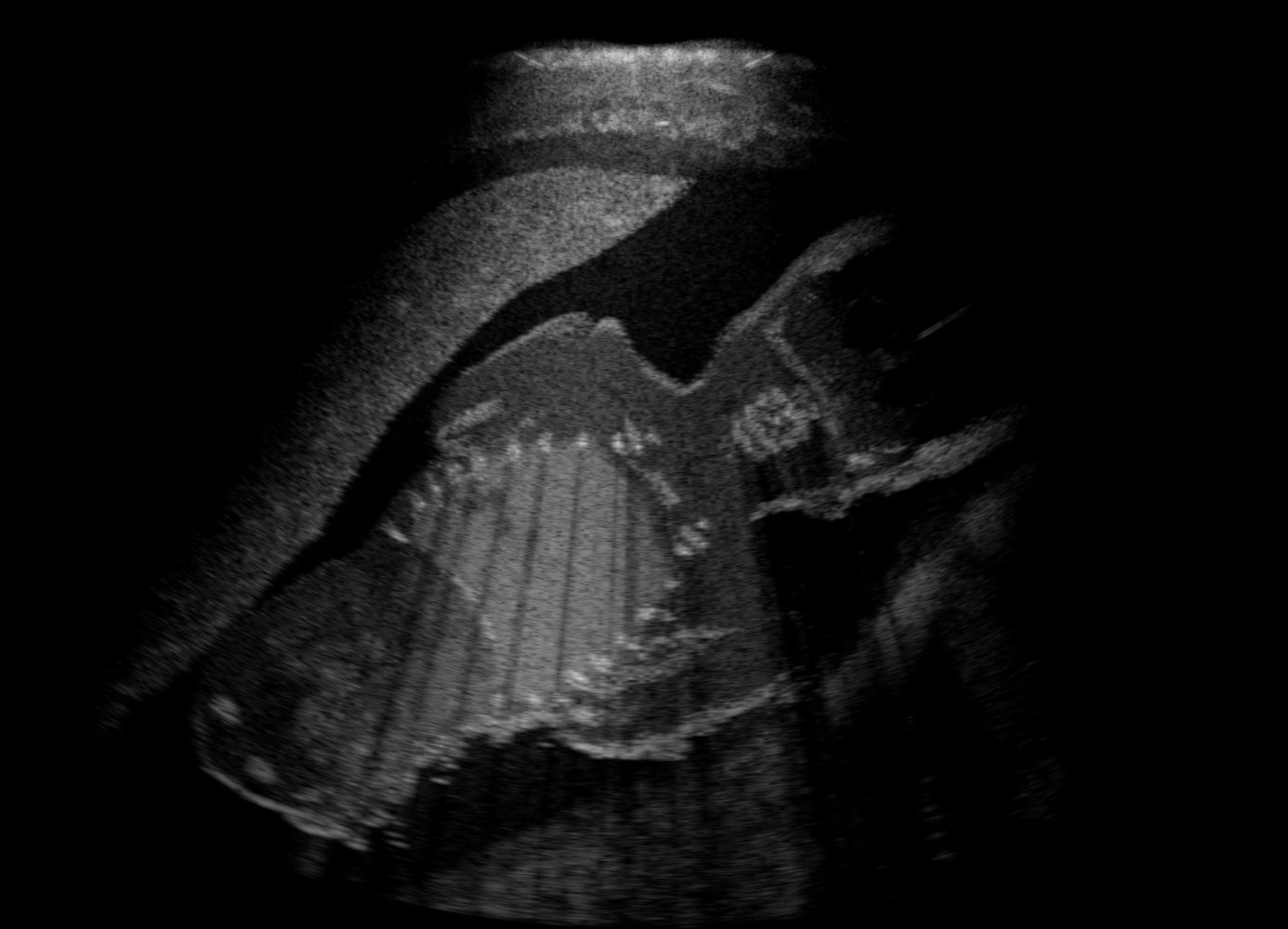}
 \includegraphics[width=1\linewidth]{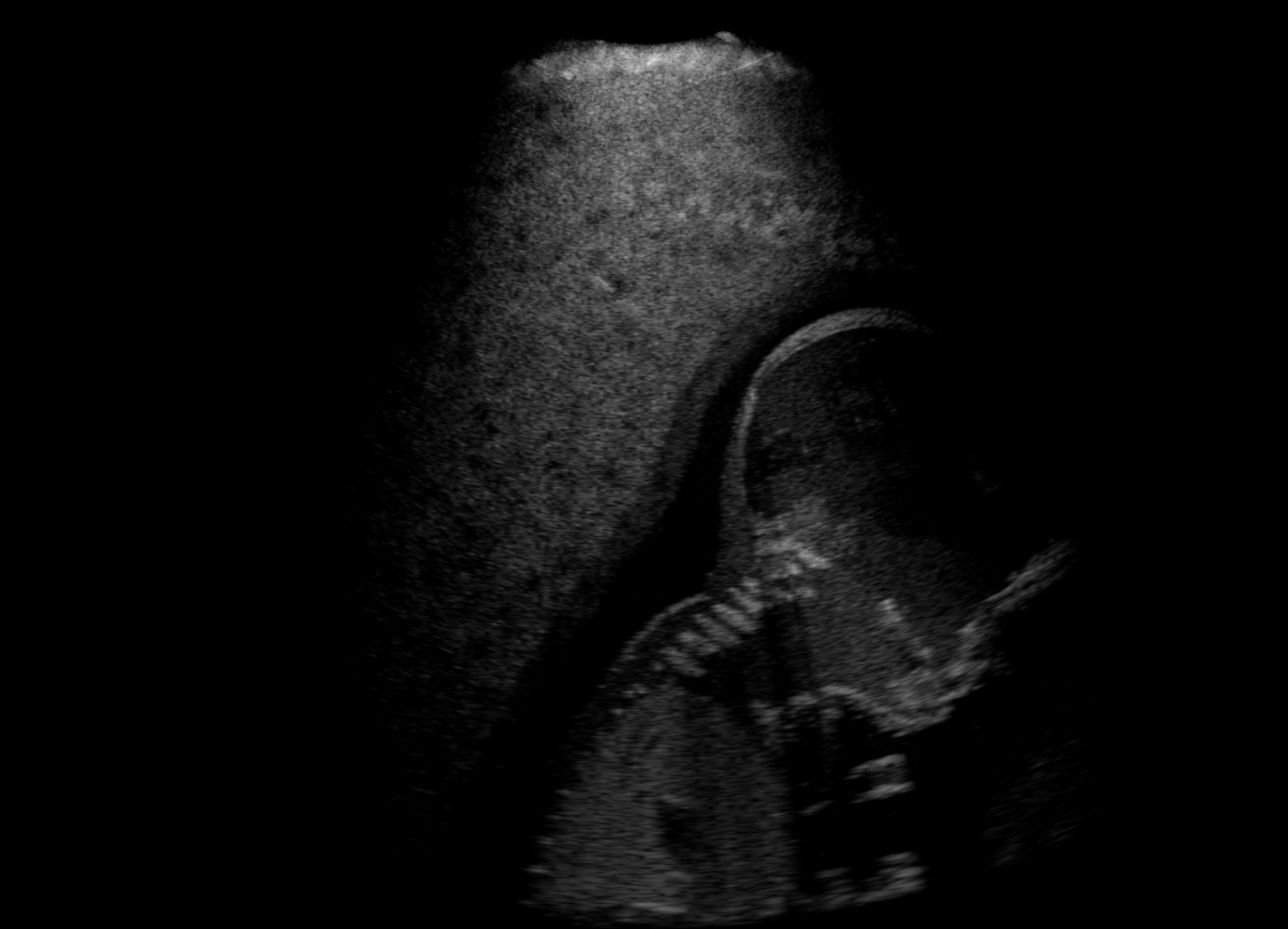}
 (e) Target
\endminipage
\caption{Inference on full field-of-view (FoV) images.} 
\label{fig:qual_results_hr}
\end{figure}

Above image translation has been demonstrated on patches.
For the entire field-of-view (FoV) US images, patch fusion from image translation of non-overlapping patches would cause artifacts at image seams. 
Averaging overlapping patches, on the other hand, would blur the essential US texture.
Although seamless tiling of US images is possible using graphical models~\cite{flach2016pure}, this requires prohibitively long computation time.
Herein, we instead directly apply our trained generator on full FoV low-quality images, since the generator is fully convolutional and thus can operate on images of arbitrary size. 
Fig.~\ref{fig:qual_results_hr} shows two examples of translated images by \textit{L2H$_\mathrm{MS}$} and \textit{L2H$_\mathrm{MSA}$}, demonstrating direct inference on full FoV images. 
While anatomical structures are well preserved and the effect of attenuation integral map is apparent, speckle texture appearance is seen to degrade slightly especially in the top image regions, where the ultrasound texture looking particularly different due to focusing difference and near-field effects.

\section{Discussion and Conclusions}
We have proposed a patch based generative adversarial network for improving the quality of simulated US images, via image translation of computationally low-cost images to high quality simulation outputs. 
Providing segmentation and attenuation integral maps to the translation framework greatly improves preservation of anatomical structures and synthesis of important acoustic shadows. 
Continuous simulation parameters, such as transmit focus and depth-dependent lateral resolution, are implicitly captured by our framework, thanks to training on image patches.
For discrete simulation parameters such as imaging mode and transducer frequency that can take a handful of different values in typical clinical imaging, it is feasible to train a separate GAN for each such setting.

Rendering high and low quality images takes 75\,ms and 40\,ms, respectively, with segmentation and attenuation maps computable within the same rendering pass. Our network inference time with a non-optimized code is 12.6\,ms on average for full FoV images on a GTX 2080\,Ti using TensorRT.
This timing improvement is rather a lower-bound, since network inference can be further accelerated, e.g.\ with FPGAs~\cite{guo2017survey}.
Furthermore, since a pass through the network runs in constant time, potential time gain can be arbitrarily high depending on the desired complexity of the target US simulation.
Although the convolutional network can process arbitrary sized image, translating full FoV images without any artifacts is still a challenge.\\
Funding for this work was provided by the Swiss Innovation Agency Innosuisse.

\newpage
\bibliographystyle{splncs04}
\bibliography{image_translation_us}

\end{document}